\documentclass[runningheads]{llncs}
\usepackage{times}  
\usepackage{helvet}  
\usepackage{courier}  
\usepackage{url}  
\usepackage{graphicx} 

\usepackage{amsfonts}
\usepackage{amsmath,amssymb}
\usepackage{isomath}
\usepackage{latexsym}
\usepackage{epic}
\usepackage{epsfig}
\usepackage{hyperref}
\usepackage{multirow}
\usepackage{hhline}
\usepackage{balance}
\usepackage{wrapfig}
\usepackage{verbatim}
\usepackage[justification=centering]{caption}
\usepackage{enumitem}
\usepackage{balance}
\usepackage{color}
\allowdisplaybreaks[1]
\usepackage{setspace}
\usepackage{thmtools}
\usepackage{thm-restate}
\usepackage[ruled,vlined, linesnumbered, boxed]{algorithm2e}
\usepackage[labelformat=simple]{subcaption}
\usepackage{balance}

\usepackage{cleveref}
\usepackage{footmisc}

\newcommand{\tocite}[1]{{\color{red} (cite!!)}}

\newcommand{\N}{\mathbb{N}}

\newcommand{\R}{\mathbb{R}}

\begin{document}
\title{Harnessing Heterogeneity: Learning from Decomposed Feedback in Bayesian Modeling}
%
%
\author{Kai Wang \and
Bryan Wilder \and
Sze-chuan Suen \and
Bistra Dilkina \and
Milind Tambe
}
\authorrunning{K. Wang et al.}
%
\institute{
Harvard University, USA \\
\email{\{kaiwang, bwilder\}@g.harvard.edu, milind\_tambe@harvard.edu} \\
University of Southern California, USA \\
\email{\{ssuen,dilkina\}@usc.edu}
}
\maketitle              
\begin{abstract}
There is significant interest in learning and optimizing a complex system composed of multiple sub-components, where these components may be agents or autonomous sensors. Among the rich literature on this topic, agent-based and domain-specific simulations can capture complex dynamics and subgroup interaction, but optimizing over such simulations can be computationally and algorithmically challenging. Bayesian approaches, such as Gaussian processes (GPs), can be used to learn a computationally tractable approximation to the underlying dynamics but typically neglect the detailed information about subgroups in the complicated system. We attempt to find the best of both worlds by proposing the idea of \textit{decomposed feedback}, which captures group-based heterogeneity and dynamics. We introduce a novel \textit{decomposed GP regression} to incorporate the subgroup decomposed feedback. Our modified regression has provably lower variance -- and thus a more accurate posterior -- compared to previous approaches; it also allows us to introduce a \textit{decomposed GP-UCB} optimization algorithm that leverages subgroup feedback. The Bayesian nature of our method makes the optimization algorithm trackable with a theoretical guarantee on convergence and no-regret property. To demonstrate the wide applicability of this work, we execute our algorithm on two disparate social problems: infectious disease control in a heterogeneous population and allocation of distributed weather sensors. Experimental results show that our new method provides significant improvement compared to the state-of-the-art.

\keywords{Decomposed feedback \and Decomposed GP regression \and D-GPUCB}
\end{abstract}
\section{Introduction}


Many complex systems involve heterogeneous subgroup behaviors which interact in physical or social structures. Analysis of these systems often exhibits a tension exemplified by two commonly used techniques: \textit{simulation} and \textit{optimization}. Simulations can capture more system dynamics, increasing the accuracy and realism of the model. For example, a precise disease and vaccination model \cite{woolthuis2017variation,del2013mathematical} can help the government predict the future effect of a disease prevention policy.
However, it is often difficult to find optimal decisions with respect to such models. Policymakers may wish to optimally allocate resources to control a disease outbreak \cite{vynnycky2008estimating}, but it is computationally and algorithmically challenging to optimize based on simulations.
Simulation-based optimization techniques have also been developed \cite{nguyen2014review} but typically lack theoretical guarantees.
On the other hand, Bayesian approaches, such as Gaussian processes (GPs), can be used to learn computationally tractable approximations of the underlying system, to which researchers can directly apply existing optimization techniques. The adaptiveness of Bayesian approaches, especially in geolocation \cite{ouyang2014multi}, makes them very popular in the weather prediction \cite{chen2014wind}.
However Bayesian approaches typically use black-box models which ignore subgroup information.
The lack of subgroup structure makes such Bayesian approaches less suitable for complex problems.

In this paper, we aim to capture subgroup heterogeneity of complex systems in Bayesian approaches through \textit{decomposed feedback}.
We formulate the underlying system as an unknown noisy black-box function, which we can only learn about from prior samples.
For example, many diseases, such as the flu, have annual cycles and other structure that characterizes its underlying dynamics and can only be learned through many years of observation.
We study such problems from the perspective of online learning, where a decision maker aims to optimize an unknown objective function \cite{brochu2010tutorial}.
At each step, the decision maker commits to an action and receives the objective value for that action.
For instance, a policymaker may implement a disease control policy \cite{sah2018optimizing} for a given time period and observe the number of subsequent infections.
This information allows the decision maker to update their knowledge of the unknown function.
The goal is to obtain low cumulative regret, which measures the difference in objective value between the actions that were taken and the true (unknown) optimum.

This problem has been well-studied in optimization and machine learning. When a parametric form is not available for the objective (as is often the case with complex systems that are difficult to model analytically), a common approach uses a GP as a nonparametric prior over smooth functions.
This Bayesian approach allows the decision maker to form a posterior distribution over the unknown function's values.
Consequently, the GP-UCB algorithm, which iteratively selects the point with the highest upper confidence bound according to the posterior, achieves a no-regret guarantee \cite{srinivas2009gaussian}.

While GP-UCB and similar techniques \cite{contal2014gaussian,wang2016optimization} have enjoyed a great deal of interest in the pure black-box setting, many physical or social systems naturally admit an intermediate level of feedback.
This may occur when the system is composed of multiple interacting sub-components or sub-groups, each of which can be measured individually or have known relationships. This occurs fairly commonly in complex systems where we want to optimize some population outcome -- individuals may naturally be grouped into sub-populations with observable outcomes over time.
For instance, disease spread, such as flu, in a population is a product of interactions between individuals in different demographic groups or locations \cite{woolthuis2017variation}.
Policymakers often have access to estimates of the prevalence of infected individuals within each subgroup \cite{del2013mathematical,wilder132018preventing} while aiming to learn and minimize the total sum of infections across subgroups.
Similarly, climate systems involve the interaction of many different variables \cite{an2011agent} (heat, wind, humidity, etc.) which can be sensed individually then combined in a nonlinear fashion to produce outputs of interest (e.g., an individual's risk of heat stroke) \cite{staiger2012perceived}.

In this paper, we demonstrate how individual heterogeneity can be handled in a Bayesian framework, allowing us to exploit heretofore unused subgroup structure information in Bayesian approaches. This novel approach provides accurate approximations and is amenable to optimization.
Specifically, we focus on the online learning perspective with decomposed feedback.
To our knowledge, no prior work studies the challenge of integrating such decomposed feedback in online decision making.

Our first contribution is to remedy this gap by proposing a \emph{decomposed GP regression} and corresponding \emph{decomposed GP-UCB} algorithm (D-GPUCB).
D-GPUCB uses a separate GP to model each individual measurable quantity and then combines the estimates to produce a posterior over the final objective.
Our second contribution is a theoretical no-regret guarantee for D-GPUCB, ensuring that its decisions are asymptotically optimal.
Third, we prove that the posterior variance at each step must be less than the posterior variance of directly using a GP to model the final objective.
This formally demonstrates that more detailed modeling reduces predictive uncertainty.
Finally, we conduct experiments in two domains using real-world data: flu prevention and heat sensing.
In each case, D-GPUCB achieves substantially lower cumulative regret than previous approaches.

\subsection{Additional Related Works}
In addition to the prior work discussed previously, there is a body of existing literature on decomposition methods \cite{brown2012multi}. For instance, the advantages of function decomposition for optimization have been discussed by \cite{khamisov2017objective}.
They demonstrated the use of decomposition in general optimization problems, but they did not study the online learning case nor provide theoretical guarantees on the convergence rate.

In the online learning setting, researchers also studied the benefits of using additive models \cite{kandasamy2015high,rolland2018high,hoang2017decentralized}.
The most related work is probably \cite{hoang2018decentralized}.
They studied the benefit of linearly additive structure even with possibly overlapping lower-dimensional inputs.
But their method cannot handle non-linear structures, which compromises its applicability.
They also did not consider decomposed feedback.
Therefore the regret bound derived from their approach is different from what we find in this paper.

In applications such as flu prevention, the underlying system dynamics often depend on the full interactions between subgroups and are not separable into lower-dimensional components.
Therefore, we consider the general setting where the subcomponents are full-dimensional and may be composed nonlinearly to produce the target (as shown in Section \ref{sec:function_decomposition}).
Both full-dimensionality and nonlinearity are not addressed in the previous literature.
This general perspective is necessary to capture common policy settings which may involve intermediate observables from simulation or domain knowledge.

\section{Preliminaries}

\subsection{Noisy Black-box Function}
Given an unknown black-box function $f: \mathcal{X} \rightarrow \R$ where $\mathcal{X} \subset \R^n$, a learner is able to select an input $\boldsymbol{x} \in \mathcal{X}$ and access the function to see the outcome $f(\boldsymbol{x})$ -- this encompasses one evaluation.
Gaussian process (GP) regression \cite{rasmussen2004gaussian} is a non-parametric method to learn the target function using Bayesian methods \cite{jones1998efficient,snoek2012practical}.
It assumes that the target function is an outcome of a GP with given kernel $k(\boldsymbol{x},\boldsymbol{x}')$ (covariance function).
GP regression is commonly used and only requires an assumption on the function smoothness.
Moreover, GP regression can handle observation error. It allows the observation at point $\boldsymbol{x}$ to be noisy: $y = f(\boldsymbol{x}) + \epsilon$, where $\epsilon \sim N(0, \sigma^2 I)$.

\subsection{Gaussian Process Regression}
Although GP regression does not require rigid parametric assumptions, a certain degree of smoothness is still needed. 
We can model $f$ as a sample from a GP: a collection of random variables, one for each $\boldsymbol{x} \in \mathcal{X}$.
A GP$(\mu(\boldsymbol{x}), k(\boldsymbol{x},\boldsymbol{x}'))$ is specified by its mean function $\mu(x) = E[f(\boldsymbol{x})]$ and covariance function $k(\boldsymbol{x},\boldsymbol{x}') = E[(f(\boldsymbol{x}) - \mu(\boldsymbol{x}))(f(\boldsymbol{x}') - \mu(\boldsymbol{x}'))]$.
For GPs not conditioned on any prior, we usually assume $\mu(\boldsymbol{x}) \equiv 0$ and bounded variance: $k(\boldsymbol{x},\boldsymbol{x}') \leq 1$.
This covariance function encodes the smoothness condition of the target function $f$ drawn from the GP.
GP regression is widely studied and used \cite{rasmussen2004gaussian,jones1998efficient,snoek2012practical}.


For a noisy sample $\boldsymbol{y}_T = [y_1, ..., y_T]^\top$ at points $A_T = \{ \boldsymbol{x}_t \}_{t \in [T]}$, $y_t = f(\boldsymbol{x}_t) + \epsilon_t ~\forall t \in [T]$ where $\epsilon_t \sim N(0, \sigma^2(\boldsymbol{x}_t))$ is Gaussian noise with variance $\sigma^2(\boldsymbol{x}_t)$. The posterior over $f$ is still a GP with posterior mean $\mu_T(\boldsymbol{x})$, covariance $k_T(\boldsymbol{x},\boldsymbol{x}')$ and variance $\sigma^2_T(\boldsymbol{x})$:
\begin{align}
\mu_T(\boldsymbol{x}) &= \boldsymbol{k}_T(\boldsymbol{x})^\top \boldsymbol{K}_T^{-1} \boldsymbol{k}_T(\boldsymbol{x}'), \label{eqn:gpr_mean}\\
k_T(\boldsymbol{x}, \boldsymbol{x}') &= k(\boldsymbol{x}, \boldsymbol{x}') - \boldsymbol{k}_T(\boldsymbol{x})^\top \boldsymbol{K}_T^{-1} \boldsymbol{k}(\boldsymbol{x}'), \label{eqn:gpr_covar} \\
\sigma^2_T(\boldsymbol{x}) &= k_T(\boldsymbol{x},\boldsymbol{x}') \label{eqn:gpr_var} 
\end{align}
where $\boldsymbol{k}_T(\boldsymbol{x}) = [k(\boldsymbol{x}_1, \boldsymbol{x}),...,k(\boldsymbol{x}_T, \boldsymbol{x})]^\top$, and $\boldsymbol{K}_T$ is the positive definite kernel matrix $[k(\boldsymbol{x}, \boldsymbol{x}')]_{\boldsymbol{x},\boldsymbol{x}' \in A_T} + \text{diag}([\sigma^2(\boldsymbol{x}_t)]_{t \in [T]})$.

\section{Decomposition}
\subsection{Function Decomposition}\label{sec:function_decomposition}
In this paper, we consider a modification to the idea of GP regression.
Suppose we have some prior knowledge of the unknown reward function $f(\boldsymbol{x})$ such that we can write the unknown function as a combination of known and unknown subfunctions: 
\begin{definition}[Linear Decomposition]\label{def:decomposition}
\begin{equation}\label{eqn:decomposition}
f(\boldsymbol{x}) = \sum\nolimits_{j=1}^J g_j(\boldsymbol{x}) f_j(\boldsymbol{x})
\end{equation}
\end{definition}
Here $g_j(\boldsymbol{x})$ are known, deterministic functions, but $f_j(\boldsymbol{x})$ are unknown functions that generate noisy observations.
For example, in the flu prevention case, the total infected population can be written as the summation of the infected population at each age \cite{del2013mathematical}.
Given treatment policy $\boldsymbol{x}$, where $\boldsymbol{x}_j$ denotes the extent to vaccinate the infected people in age group $j$, due to the complicated interactions between subgroups, the infection outcome of age group $j$ is also affected by the vaccination of all the other subgroups.
Therefore, we use a full-dimensional function $f_j(\boldsymbol{x})$ to represent the unknown infected population
at age group $j$ with its known, deterministic weighted function $g_j(\boldsymbol{x}) = 1$.
Therefore, the total infected population $f(\boldsymbol{x})$ can be simply expressed as $\sum\nolimits_{j=1}^J f_j(\boldsymbol{x})$.

Interestingly, any deterministic linear composition of outcomes of GPs still yields a GP outcome.
That means if all of the $f_j$ are generated from GPs, then the entire function $f$ can be written as an outcome of another GP \cite{rasmussen2004gaussian}.
However, this works only for the linear case.
In general, an arbitrary composition of outcomes of GPs is no longer a GP outcome.
This naturally brings up a generalization to the non-linear case:
\begin{definition}[General Decomposition]\label{def:general_decomposition}
\begin{equation}\label{eqn:general_decomposition}
f(\boldsymbol{x}) = g(f_1(\boldsymbol{x}), f_2(\boldsymbol{x}), ..., f_J(\boldsymbol{x}))
\end{equation}
\end{definition}
The function $g$ can be any deterministic function\footnote{
We assume that the coefficients of $g$ are deterministic and its value only depends on $f_i(\boldsymbol{x})$ but not $\boldsymbol{x}$ directly.
For functions $g$ with $\boldsymbol{x}$-dependent coefficients, we can encode $n = \text{dim}(x)$ more inputs with
$f(\boldsymbol{x}) = g(f_1(\boldsymbol{x}), f_2(\boldsymbol{x}),..., f_{J+n}(\boldsymbol{x}))$,
where $f_{J+i}(\boldsymbol{x}) = \boldsymbol{x}_i$.
So an $\boldsymbol{x}$-dependent function $g$ is just a subcase of Definition \ref{def:general_decomposition}.
} (e.g., polynomial, neural network).
We will cover the result of linear decomposition first and then generalize our insights to the general decomposition case.

\subsection{Bandit Problem with Decomposed Feedback}
We assume the output value of the target function is the learner's reward (penalty). The goal is then to learn the unknown underlying function $f$ while optimizing the cumulative reward.
This is usually known as an online learning or multi-arm bandit problem \cite{auer2002using,waniek2016repeated,libin2017efficient}.
In this paper, given the knowledge of decomposition function $g$ (Definition \ref{def:decomposition} or Definition \ref{def:general_decomposition}), the learner is going to repeatedly choose an input $\boldsymbol{x}_t \in \mathcal{X}$, where $t$ is the current round, and observe the value of each unknown decomposed function $f_j$ perturbed by a noise: $y_{j,t} = f_j(\boldsymbol{x}_t) + \epsilon_{j,t}, \epsilon_{j,t} \sim N(0, \sigma_j^2) ~\forall j \in [J]$.
At the same time, the learner receives the composed reward from this input $\boldsymbol{x}_t$, which is $y_t = g(y_{1,t}, y_{2,t}, ..., y_{J,t}) = f(\boldsymbol{x}_t) + \epsilon_t$ where $\epsilon_t$ is an aggregated noise.
The goal is to maximize the sum of noise-free rewards $\sum\nolimits_{t=1}^T f(\boldsymbol{x}_t)$, which is equivalent to minimizing the cumulative regret $R_T = \sum\nolimits_{t=1}^T r_t = \sum\nolimits_{t=1}^T f(\boldsymbol{x}^*) - f(\boldsymbol{x}_t)$, where $\boldsymbol{x}^* = \arg\max_{\boldsymbol{x} \in \mathcal{X}} f(\boldsymbol{x})$ and individual regret $r_t = f(\boldsymbol{x}^*) - f(\boldsymbol{x}_t)$.

This decomposed feedback is related to the semi-bandit setting, where a set of items is chosen from the ground set and feedback is received about the chosen items \cite{neu2013efficient}, which is similar to our decomposed feedback.
The major difference is that in the semi-bandit setting, the learner only selects a subset of items and receives rewards from them; in our case, the learner has to decide on an policy over all items and receive the entire reward instead of a partial reward.
Hence, the technical challenges are quite different. 
\newline

\section{Problem Statement}
Using the flu prevention as an example, a policymaker implements a yearly disease control policy and observes the number of subsequent infections.
A policy is an input $\boldsymbol{x}_t \in \mathbb{R}^n$, where each entry $x_{t,i}$ denotes the extent to vaccinate individuals in age group $i$.
For example, if the government spends more budget $x_{t,i}$ in group $i$, then the people in this group will be more likely to get a flu shot.

Given the decomposition assumption and samples (previous policies) at points $\boldsymbol{x}_t ~\forall t \in [T]$, including all the function values $f(\boldsymbol{x}_t)$ (total infected population) and decomposed function values $f_j(\boldsymbol{x}_t)$ (infected population in group $j$), the learner attempts to learn the function $f$ while simultaneously minimizing regret.
Therefore, we have two main challenges:
(i) how best to approximate the reward function using the decomposed feedback (non-parametric approximation), and (ii) how to use this estimation to most effectively reduce the average regret (bandit problem).

\subsection{Regression: Non-parametric Approximation}
Our first aim is to fully utilize the decomposed problem structure to get a better approximation of $f(\boldsymbol{x})$.
The goal is to learn the underlying disease pattern faster by using the decomposed problem structure.
Given the linear (or general) decomposition assumption and noisy samples at points $\{ \boldsymbol{x}_t\}_{t \in [T]}$, the learner can observe the outcome of each decomposed function $f_j(\boldsymbol{x}_t)$ at each sample point $\boldsymbol{x}_t ~\forall t \in [T]$.
In Section \ref{sec:decomposed_gp_regression}, we provide a Bayesian update to the unknown function which fully utilizes the learner's knowledge of the decomposition.

\subsection{Bandit Problem: Learning while Minimizing Regret}
In the flu example, each annual flu-awareness campaign is constrained by a budget, and we assume the policymaker does not know the underlying disease spread pattern.
At the beginning of each year, the policymaker chooses a new campaign policy based on the previous years' results and observes the outcome of this new policy. 
The goal is to minimize the cumulative regret (all additional infections in prior years) while learning the unknown function (disease pattern).
In Section \ref{sec:decomposed_GPUCB}, we will show how a decomposed GP regression, with a UCB-based algorithm, can be used to address these challenges.

\section{Decomposed GP Regression}\label{sec:decomposed_gp_regression}
\begin{figure}[t]
    \centering
    \begin{subfigure}[b]{0.99\textwidth}
        \includegraphics[scale=0.066]{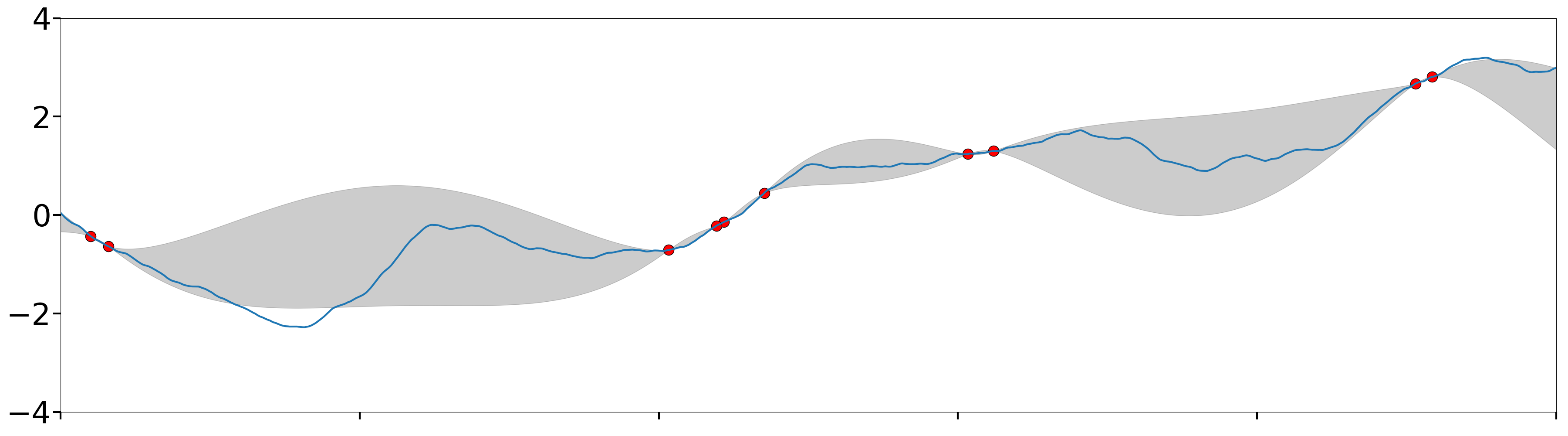}
        \hfill
        \includegraphics[scale=0.066]{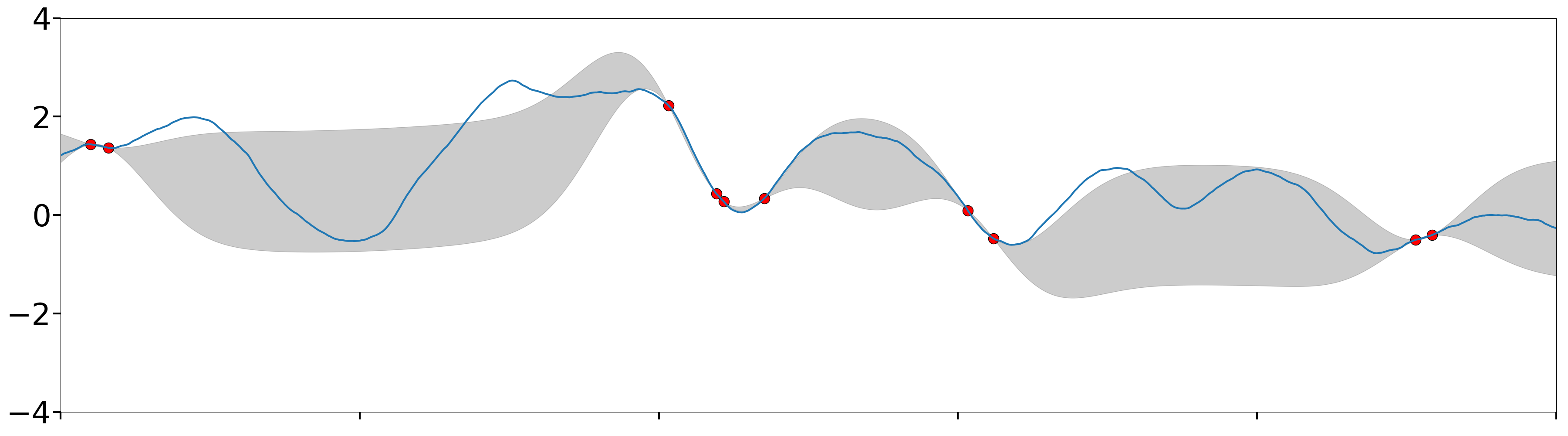}
        \vspace{-5mm}
        \caption{Decomposed functions $f_1, f_2$ and their GP regression posteriors.}
        \label{fig:decomposed_samples}
    \end{subfigure}
    \hfill
    \begin{subfigure}[b]{0.49\textwidth}
        \includegraphics[scale=0.066]{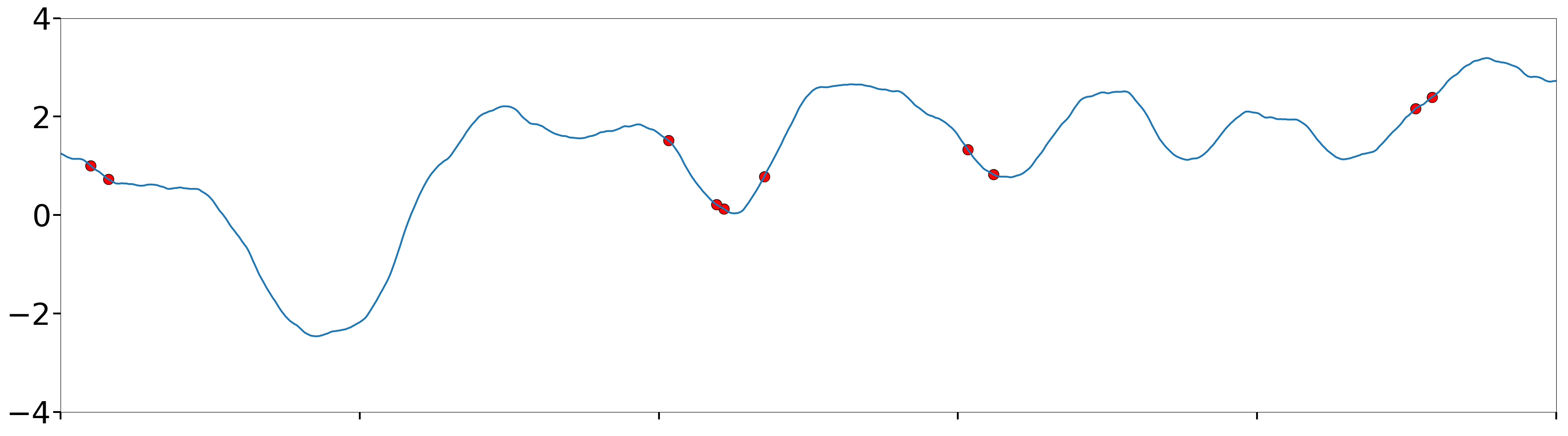}
        \vspace{-5mm}
        \caption{Entire target function $f = f_1 + f_2$ and its sampled values}
        \label{fig:prior_samples}
    \end{subfigure}
    \hfill
    \begin{subfigure}[b]{0.49\textwidth}
        \includegraphics[scale=0.066]{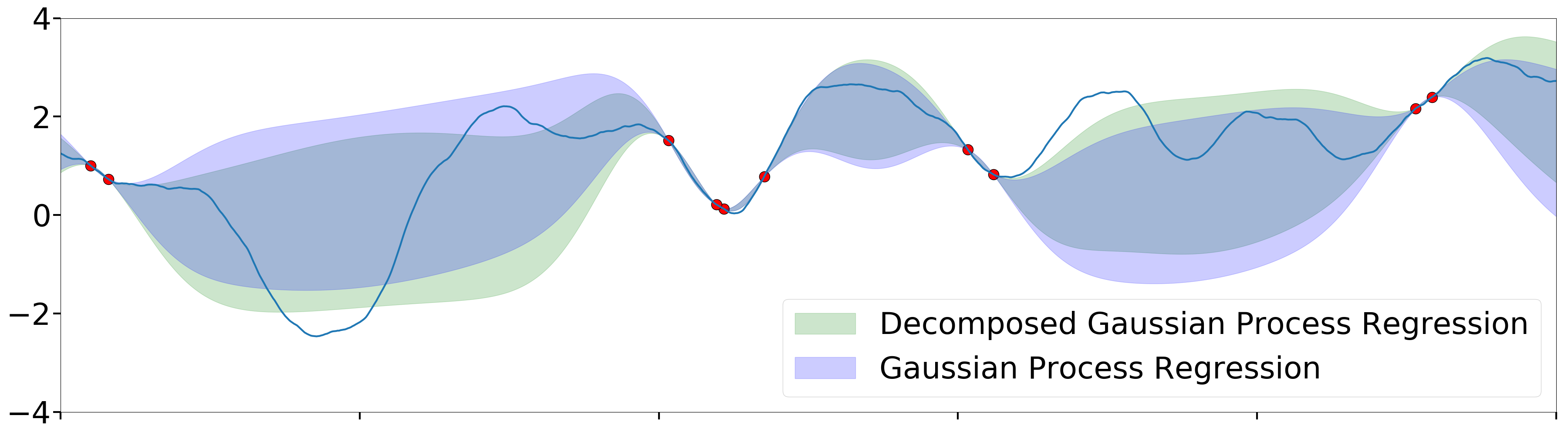}
        \vspace{-5mm}
        \caption{The posteriors of decomposed GP regression and GP regression}
        \label{fig:entire_gpr}
    \end{subfigure}
    \vspace{-3mm}
    \caption{Illustration of the comparison between decomposed GP regression (Algorithm \ref{alg:decomposed_gpr}) and standard GP regression.
    Decomposed GP regression shows a smaller average variance (0.878 v.s. 0.943) and a better estimate of the target function. 
	\label{fig:decomposed_gpr}
	}
\end{figure}

In this section, we first propose a decomposed GP regression (Algorithm \ref{alg:decomposed_gpr}), but then provide the key theoretical result (Theorem \ref{thm:reduced_variance}) in this section: decomposed GP regression provably provides a posterior with smaller variance.
Here, we mainly focus on the case of linear decomposition.

The idea behind decomposed GP regression is as follows: given the linear decomposition assumption (Definition \ref{def:decomposition}), run GP regression for each $f_j(\boldsymbol{x})$ individually and get the aggregated approximation by $f(\boldsymbol{x}) = \sum\nolimits_{j=1}^J g_j(\boldsymbol{x}) f_j(\boldsymbol{x})$ (illustrated in Figure \ref{fig:decomposed_gpr}).
Assuming we have $T$ previous samples with input $\boldsymbol{x}_1, \boldsymbol{x}_2, ..., \boldsymbol{x}_T$ and the noisy outcome of each individual function $\boldsymbol{y}_{j,t} = f_j(\boldsymbol{x}_t) + \epsilon_{j,t} ~\forall j \in [J], t \in [T]$, where $\epsilon_{j,t} \sim N(0, \sigma^2_j)$,
the outcome of the entire target function can be computed as $y_t = \sum\nolimits_{j=1}^J g_j(\boldsymbol{x}_t) y_{j,t}$.
Further assume the function $f_j(\boldsymbol{x})$ is an outcome of $GP(0, k_j) ~\forall j$.
Therefore the entire function $f$ is also an outcome of $GP(0, k)$ where $k(\boldsymbol{x}, \boldsymbol{x}') = \sum\nolimits_{j=1}^J g_j(\boldsymbol{x}) k_j(\boldsymbol{x}, \boldsymbol{x}') g_j(\boldsymbol{x}')$.

We are going to compare two ways to approximate the function $f(\boldsymbol{x})$ using existing samples: (i) directly use standard GP regression with the composed kernel $k(\boldsymbol{x},\boldsymbol{x}')$ and noisy samples $\{(\boldsymbol{x}_t, y_t)\}_{t \in [T]}$ and (ii) for each $j \in [J]$, first run standard GP regression with kernel $k_j(\boldsymbol{x},\boldsymbol{x}')$ and noisy samples $\{(\boldsymbol{x}_t, y_{j,t})\}_{t \in [T]}$, then compose the outcomes with the weighted function $g_j(\boldsymbol{x})$ to get $f(\boldsymbol{x})$. This is shown in Algorithm \ref{alg:decomposed_gpr}.

\begin{algorithm}[t]
\caption{Decomposed GP Regression}
\label{alg:decomposed_gpr}
\textbf{Input: } kernel functions $k_j(\boldsymbol{x},\boldsymbol{x}')$ to each $f_j(\boldsymbol{x})$ and previous samples $(\boldsymbol{x}_t, y_{j,t}) ~\forall j \in [J], t \in [T]$ \\
\For{$j=1,2...,J$} {
Let $\mu_{j,T}(\boldsymbol{x}), k_{j,T}(\boldsymbol{x}, \boldsymbol{x}'), \sigma^2_{j,T}(\boldsymbol{x})$ be the output of GP regression with $k_j(\boldsymbol{x},\boldsymbol{x}')$ and $(\boldsymbol{x}_t, y_{j,t})$. \\
}
\textbf{Return: } $k_T(\boldsymbol{x}, \boldsymbol{x}') = \sum\nolimits_{j=1}^J g_j(\boldsymbol{x}) k_{j,T}(\boldsymbol{x}, \boldsymbol{x}') g_j(\boldsymbol{x}')$, $\mu_T(\boldsymbol{x}) = \sum\nolimits_{j=1}^J g_j(\boldsymbol{x}) \mu_{j,T}(\boldsymbol{x})$, $\sigma_T^2(\boldsymbol{x}) = k_T(\boldsymbol{x}, \boldsymbol{x})$
\end{algorithm}

Now we turn to the key result of this section, which is to analytically compare GP regression and decomposed GP regression (Algorithm \ref{alg:decomposed_gpr}).
We will compute the variance returned by both algorithms and show that the latter variance is smaller than the former.
Missing proofs are in the Appendix for brevity (in the following, the subscripts ``entire" and ``decomp" refer to cases without and with decomposition respectively). 
\begin{restatable}[]{proposition}{entireVariance}\label{prop:entire_variance}
The variance returned by the standard GP regression is
\begin{equation}\label{eqn:entire_variance}
\sigma_{T,\text{entire}}^2(\boldsymbol{x}) = k(\boldsymbol{x}, \boldsymbol{x}) - \sum\nolimits_{i,j} \boldsymbol{z}_i^\top (\sum\nolimits_{l} \boldsymbol{D}_l \boldsymbol{K}_{l,T} \boldsymbol{D}_l)^{-1} \boldsymbol{z}_j
\end{equation}
where $\boldsymbol{D}_j \! = \! \text{diag}([g_j(\boldsymbol{x}_1), ..., g_j(\boldsymbol{x}_T)])$ and $\boldsymbol{z}_i \! = \! \boldsymbol{D}_i \boldsymbol{k}_{i,T}(\boldsymbol{x}) g_i(\boldsymbol{x}) \! \in \! \R^T$.
\end{restatable}

\begin{restatable}[]{proposition}{decomposedVariance}\label{prop:decomposed_variance}
The variance returned by Algorithm \ref{alg:decomposed_gpr} is
\begin{equation}\label{eqn:decomposed_variance}
\sigma^2_{T, \text{decomp}}(\boldsymbol{x}) = k(\boldsymbol{x},\boldsymbol{x}) - \sum\nolimits_{l} \boldsymbol{z}_l^\top (\boldsymbol{D}_l \boldsymbol{K}_{l,T} \boldsymbol{D}_l)^{-1} \boldsymbol{z}_l
\end{equation}
\end{restatable}

In order to show that our approach has lower variance, we first recall the matrix-fractional function and its convex property.
\begin{lemma}\label{lemma:matrix_fractinoal_function}
The matrix-fractional function $h(\boldsymbol{X}, \boldsymbol{y}) \! = \! \boldsymbol{y}^\top \boldsymbol{X}^{-1} \boldsymbol{y}$ is defined and also convex on $\text{dom} f \! = \! \{ (\boldsymbol{X}, \boldsymbol{y}) \in \mathbf{S}^T_+ \times \R^T \}$.
\end{lemma}

Now we can compare the variance provided by Proposition \ref{prop:entire_variance} and Proposition \ref{prop:decomposed_variance}.

\begin{restatable}[]{theorem}{reducedVariance}\label{thm:reduced_variance}
The variance provided by decomposed GP regression (Algorithm \ref{alg:decomposed_gpr}) is less than or equal to the variance provided by GP regression, which implies the uncertainty by using decomposed GP regression is smaller.
\end{restatable}



Theorem \ref{thm:reduced_variance} implies that decomposed GP regression provides a posterior with smaller variance, which could be considered as the uncertainty of the approximation.
Since the posterior belief after the GP regression is still a GP, a smaller variance directly implies a more concentrated Gaussian distribution, leading to less uncertainty and smaller root-mean-squared error. 
Intuitively, this is because Algorithm \ref{alg:decomposed_gpr} exploits the knowledge of decomposition but the standard GP regression does not.
This idea of handling decomposition in the Bayesian context is very general and can be applied to many problems.

In the case of general decomposition, unfortunately, a non-linear composition of GPs may not be a GP, so we cannot guarantee function $f$ to be an outcome of a GP, and it is difficult to provide guarantees of lower variance.
Nonetheless we address this general case later in Section \ref{sec:generalized_DGPUCB}.


\section{Decomposed GP-UCB Algorithm}\label{sec:decomposed_GPUCB}
The goal of a traditional bandit problem is to optimize the objective function $f(\boldsymbol{x})$ by minimizing the regret.
In Section \ref{sec:information_gain}, we briefly introduce some prior work that do not use decomposed feedback.
In Section \ref{sec:DGPUCB}, we introduce our novel algorithm, D-GPUCB, with corresponding theoretical guarantee and its comparison with the previous work.
In Section \ref{sec:generalized_DGPUCB}, we extend our algorithm to the case of general decomposition, which is not addressed by the previous GP-UCB algorithm.
To summarize our contribution in this paper, our algorithm D-GPUCB provides the following advantages:
(i) D-GPUCB applies decomposed GP regression with decomposed feedback knowledge, which incurs lower variance at each step (Theorem \ref{thm:reduced_variance}) compared to normal GP regression.
(ii) The regret bound provided by D-GPUCB is easy to compute, as it only requires knowledge on individual subfunctions but not the entire complex composed function (Theorem \ref{thm:decomposed_GPUCB_bound}).
(iii) D-GPUCB and its theoretical guarantee generalize to the non-linear composition of GPs (Theorem \ref{thm:generalized_decomposed_GPUCB_bound}), which previous work does not handle.
We will explain these contributions in detail in the following sections.

\subsection{GP-UCB and Information Gain}\label{sec:information_gain}

The GP-UCB algorithm for bandit optimization was introduced by \cite{srinivas2009gaussian}.
GP-UCB explores a target function $f$, an outcome of GP with kernel $k(\boldsymbol{x}, \boldsymbol{x}')$, by balancing exploration and exploitation. Specifically, at each step the algorithm selects the point $x$ for which the random variable $f(x)$ has highest upper confidence bound, calculated as the weighted sum of the posterior mean and variance. Once a new value is observed, the algorithm performs a Bayesian update using GP regression. 

It would seem at first glance that we could compute the regret bound for our case of decomposed feedback in the following way:
given the linear decomposition (Definition \ref{def:decomposition}) and the additive property of GPs \cite{rasmussen2004gaussian}, the entire function $f(\boldsymbol{x})$ is still an outcome of a GP with a linearly composed kernel $k(\boldsymbol{x},\boldsymbol{x}') = \sum\nolimits_{j=1}^J g_j(\boldsymbol{x}) k_j(\boldsymbol{x},\boldsymbol{x}') g_j(\boldsymbol{x}')$, where $k(\boldsymbol{x},\boldsymbol{x}') \leq \sum\nolimits_{j=1}^J B_j^2 = B^2$ with $B_j = \max\nolimits_{\boldsymbol{x} \in \mathcal{X}} |g_j(\boldsymbol{x})|$.
Then we can apply the regret bound of GP-UCB and get a similar bound:
\begin{align}\label{eqn:GPUCB_regret_bound}
Pr\{ R_T \leq \sqrt[]{C_1 T \beta_T B^2 \gamma_{\text{entire}, T}} ~\forall T \geq 1 \} \geq 1 - \delta
\end{align}
where $\beta_T$ are appropriate constants. The information gain $\gamma_{\text{entire}, T}$ is a kernel-specific quantity used in Srinivas et al.'s analysis. For any given kernel, $\gamma_T$ should upper bound the mutual information $I(y_A;f_A)$ for any set of points $A$ with $|A| = T$, where $y_A = f_A + \epsilon_A$ are the noisy function values.


Unfortunately, we cannot compute the bound given by inequality \eqref{eqn:GPUCB_regret_bound} easily, because it depends on  $\gamma_{\text{entire}, T}$, which is difficult to compute for the \emph{composed} kernel $k$.
In general, $\gamma_{\text{entire}, T}$ is \emph{not} a simple function of the underlying kernels $k_j$.
This is a serious drawback, indicating that there is little principled basis for using GP-UCB in complex systems with underlying heterogeneity.
Therefore, a natural question arises: can we do better with the decomposed feedback from each individual component?

\subsection{Decomposed GP-UCB Algorithm and No-Regret Property Property}\label{sec:DGPUCB}
In our bandit problem with decomposed feedback, given the knowledge of linear decomposition and decomposed feedback, the learner is able to access samples of individual functions $f_j(\boldsymbol{x})$.
In this section, we turn to determining how we can apply our decomposed GP regression (Algorithm \ref{alg:decomposed_gpr}) to achieve a better regret bound.
This leads to our second contribution: the decomposed GP-UCB algorithm, which uses decomposed GP regression when decomposed feedback is accessible.

\begin{algorithm}[t]
\caption{Decomposed GP-UCB}
\label{alg:decomposed_gpucb}
\textbf{Input: } Input space $\mathcal{X}$; GP priors $\mu_{j,0}, \sigma_{j,0}, k_j ~\forall j \in [J]$ \\
\For{t = 1,2,...} {
    Compute all mean $\mu_{j,t-1}$ and variance $\sigma^2_{j,t-1} \forall j$\\
    $\mu_{t-1}(\boldsymbol{x}) = \sum\nolimits_{j=1}^J g_j(\boldsymbol{x}) \mu_{j,t-1}(\boldsymbol{x})$, \quad $\sigma_{t-1}^2(\boldsymbol{x}) = \sum\nolimits_{j=1}^J g_j^2(\boldsymbol{x}) \sigma_{j,t-1}^2$ \\
    Choose $\boldsymbol{x}_t = \arg\max_{\boldsymbol{x} \in \mathcal{X}} \mu_{t-1}(\boldsymbol{x}) + \sqrt[]{\beta_t} \sigma_{t-1}(\boldsymbol{x}) $ \\
    Sample $y_{j,t} = f_j(\boldsymbol{x}_t) ~\forall j\!\in\![J]$ \\
    Perform Bayesian update to get $\mu_{j,t}, \sigma_{j,t} ~\forall j\!\in\![J]$
}
\end{algorithm}

Algorithm \ref{alg:decomposed_gpucb} performs Bayesian update (line 8) with our decomposed GP regression (Algorithm \ref{alg:decomposed_gpr} and line 4-5).
According to Theorem \ref{thm:reduced_variance}, our algorithm takes advantage of decomposed feedback and provides a more accurate and less uncertain approximation at each iteration.
This suggests that D-GPUCB may perform better than GP-UCB by adopting decomposed feedback, which will be shown in the experiments.
Here we provide a new regret bound on our D-GPUCB in Theorem \ref{thm:decomposed_GPUCB_bound}.

\begin{restatable}[]{theorem}{decomposedGPUCBbound}\label{thm:decomposed_GPUCB_bound}
Let $\delta \in  (0,1)$ and $\beta_t = 2 \log(|\mathcal{X}| t^2 \pi^2 / 6 \delta )$.
Running decomposed GP-UCB (Algorithm \ref{alg:decomposed_gpucb}) for a composed sample $f(\boldsymbol{x}) = \sum\nolimits_{j=1} g_j(\boldsymbol{x}) f_j(\boldsymbol{x})$ with bounded variance $k_j(\boldsymbol{x},\boldsymbol{x}) \leq 1$ and each $f_j \sim GP(0, k_j(\boldsymbol{x},\boldsymbol{x}'))$,
we obtain a regret bound of $\mathcal{O}(\sqrt[]{T \log |\mathcal{X}| \sum_{j=1}^J B_j^2 \gamma_{j,T} })$ with high probability $1-\delta$, where $B_j = \max\nolimits_{\boldsymbol{x} \in \mathcal{X}} |g_j(\boldsymbol{x})|$ and $\gamma_{j,T}$ is the maximum information gain of function $f_j$.
Precisely,
\begin{equation}\label{eqn:decomposed_gpucb_bound}
\text{Pr}\big\{ R_T \leq \sqrt[]{C_1 T \beta_T \sum\nolimits_{j=1}^J B_j^2 \gamma_{j,T} } ~\forall T \! \geq \! 1 \big\} \geq 1 - \delta
\end{equation}
where $C_1 = 8/ \log (1 + \sigma^{-2})$ with noise variance $\sigma^{2}$. 
\end{restatable}

Notice that the regret bound given by inequality \eqref{eqn:decomposed_gpucb_bound} only requires the knowledge of $\gamma_{j,T}$ (instead of $\gamma_{\text{entire}, T}$).
Given the knowledge of individual kernel $k_j$ and upper bounds provided by \cite{srinivas2009gaussian} on the information gain, each $\gamma_{j,T}$ can be easily computed.
Instead, the regret bound given by GP-UCB (inequality \ref{eqn:GPUCB_regret_bound}) requires an upper bound on the composed kernel, which as mentioned in Section \ref{sec:information_gain}, is difficult to compute.

\subsection{Generalized Decomposed GP-UCB Algorithm and No-Regret Property}\label{sec:generalized_DGPUCB}

We now consider the general decomposition (Definition \ref{def:general_decomposition}): $f(\boldsymbol{x}) = g(f_1(\boldsymbol{x}), f_2(\boldsymbol{x}), ..., f_J(\boldsymbol{x}))$.
To achieve the no-regret property, we further require the function $g$ to have bounded partial derivatives \sloppy $| \nabla_j g(\boldsymbol{x}) | \leq B_j ~\forall j \in [J]$.
This corresponds to the linear decomposition case, where $| \nabla_j g | = | g_j(\boldsymbol{x}) | \leq B_j$.
Since a non-linear composition of GPs is no longer a GP, the standard GP-UCB algorithm does not have any guarantees for this setting. However, we show that our approach, which leverages the special structure of the problem, still enjoys a no-regret guarantee:

\begin{algorithm}[t]
\caption{Generalized Decomposed GP-UCB}
\label{alg:generalized_decomposed_gpucb}
\textbf{Input: } Input space $\mathcal{X}$; GP priors $\mu_{j,0}, \sigma_{j,0}, k_j ~\forall j \in [J]$ \\
\For{t = 1,2,...} {
    Compute the aggregated mean and variance bound: \\
    $\mu_{t-1}(\boldsymbol{x}) = g(\mu_{1,t-1}(\boldsymbol{x}), ..., \mu_{J,t-1}(\boldsymbol{x}) )$, \quad $\sigma_{t-1}^2(\boldsymbol{x}) = J \sum\nolimits_{j=1}^J B_j^2 \sigma_{j,t-1}^2(\boldsymbol{x})$ \\
    Choose $\boldsymbol{x}_t = \arg\max_{\boldsymbol{x} \in \mathcal{X}} \mu_{t-1}(\boldsymbol{x}) + \sqrt[]{\beta_t} \sigma_{t-1}(\boldsymbol{x}) $ \\
    Sample $y_{j,t} = f_j(\boldsymbol{x}_t) ~\forall j\!\in\![J]$ \\
    Perform Bayesian update to get $\mu_{j,t}, \sigma_{j,t} ~\forall j\!\in\![J]$
}
\end{algorithm}

\begin{restatable}[]{theorem}{generalizedDecomposedGPUCBbound}\label{thm:generalized_decomposed_GPUCB_bound}
Using generalized decomposed GP-UCB with $\beta_t = 2 \log(|\mathcal{X}| J t^2 \pi^2 / 6 \delta )$ for a composed sample $f(\boldsymbol{x}) = g(f_1(\boldsymbol{x}), ..., f_J(\boldsymbol{x}))$ of GPs with bounded variance $k_j(\boldsymbol{x},\boldsymbol{x}) \leq 1$ and each $f_j \sim GP(0, k_j(\boldsymbol{x},\boldsymbol{x}'))$.
we obtain a regret bound of $\mathcal{O}(\sqrt[]{T \log |\mathcal{X}| \sum_{j=1}^J B_j^2 \gamma_{j,T} })$ with high probability, where $B_j = \max\limits_{\boldsymbol{x} \in \mathcal{X}} |\nabla_j g(\boldsymbol{x}) |$ and $\gamma_{j,T}$ is the maximum information gain of function $f_j$.
Precisely,
\begin{equation}\label{eqn:decomposed_GPUCB_regret_bound}
\text{Pr}\big\{ R_T \leq \sqrt[]{C_1 T \beta_T \sum\nolimits_{j=1}^J B_j^2 \gamma_{j,T} } ~ \forall T \geq 1 \big\} \geq 1 - \delta
\end{equation}
where $C_1 = 8/ \log (1 + \sigma^{-2})$ with noise variance $\sigma^2$. 
\end{restatable}

The intuition is that so long as each individual function is drawn from a GP, we can still perform GP regression on each function individually to get an estimate of each decomposed component.
Based on these estimates, we compute the corresponding estimate to the final objective value by combining the decomposed components with the function $g$.
Since the gradient of function $g$ is bounded, we can propagate the variance of each individual approximation to the final objective function, which allows us to get a bound on the total uncertainty.
Consequently, we can prove a high-probability bound between our algorithm's posterior distribution and the target function, which enables us to bound the cumulative regret by a similar technique as Theorem \ref{thm:decomposed_GPUCB_bound}.
Interestingly, we lose a factor $J$ in the $\beta_t$ and Algorithm \ref{alg:generalized_decomposed_gpucb} must be modified slightly from the previous algorithms due to the non-linearity.



Despite the underlying unknown function not being an outcome of GP, our D-GPUCB algorithm still works with theoretical guarantee on the regret bound.
Our result greatly enlarges the set of problems where Bayesian approaches can be applied.
We have shown that the generalized D-GPUCB preserves the no-regret property even when the underlying function is a non-linear composition of GPs. 
This showcases the contribution of our algorithm and its applicability.

\section{Continuous Sample Space}
All the above theorems are for discrete sample spaces $\mathcal{X}$.
However, many real-world scenarios have a continuous space.
\cite{srinivas2009gaussian} used the discretization technique to reduce a continuous space to a discrete case by using a larger exploration constant:
\begin{align*}
\beta_t = 2 \log(2 t^2 \pi^2 / (3 \delta)) + 2 d \log(t^2 d b r \sqrt[]{\log(4 d a / \delta)})
\end{align*}
while assuming $\text{Pr}\{ \sup_{\boldsymbol{x} \in \mathcal{X}} |\partial f / \partial \boldsymbol{x}_i | \!>\! L \} \! \leq \! a e^{-(L/b)^2}$.
(In the general decomposition case, $\beta_t = 2 \log(2 J t^2 \pi^2 / (3 \delta)) + 2 d \log(t^2 d b r \sqrt[]{\log(4 d a / \delta)})$).
All of our proofs directly follow using the same technique.
Therefore the no-regret property and regret bound (Theorem \ref{thm:decomposed_GPUCB_bound} and \ref{thm:generalized_decomposed_GPUCB_bound}) also hold in continuous sample spaces as in \cite{srinivas2009gaussian}.

\section{Experiments}
\begin{figure}[t]
    \centering
    \begin{subfigure}[b]{0.24\textwidth}
        \includegraphics[scale=0.08]{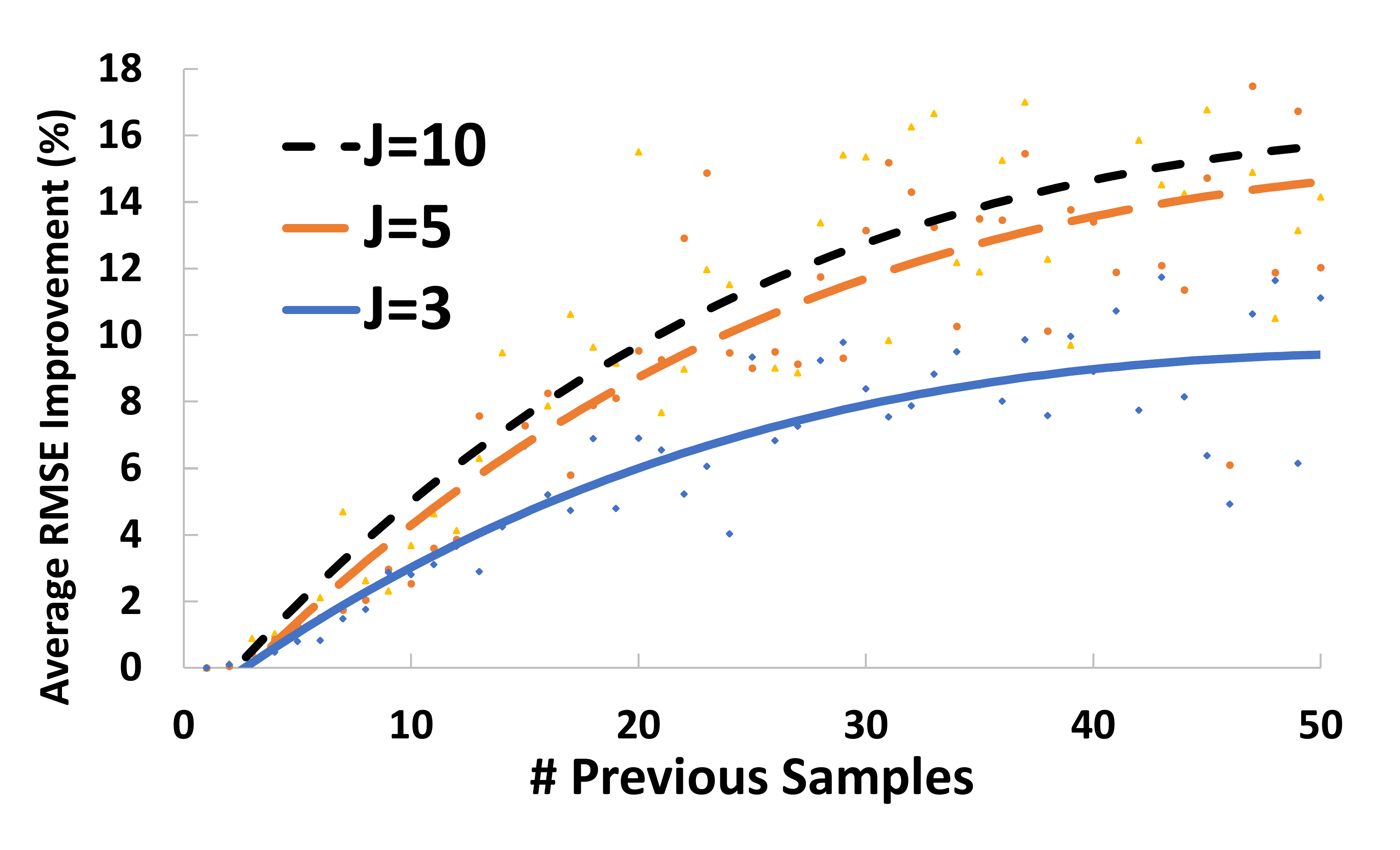}
        \vspace{-1.5mm}
        \caption{Square exponential}
        \label{fig:RBF_kernel}
    \end{subfigure}
    \begin{subfigure}[b]{0.24\textwidth}
        \includegraphics[scale=0.08]{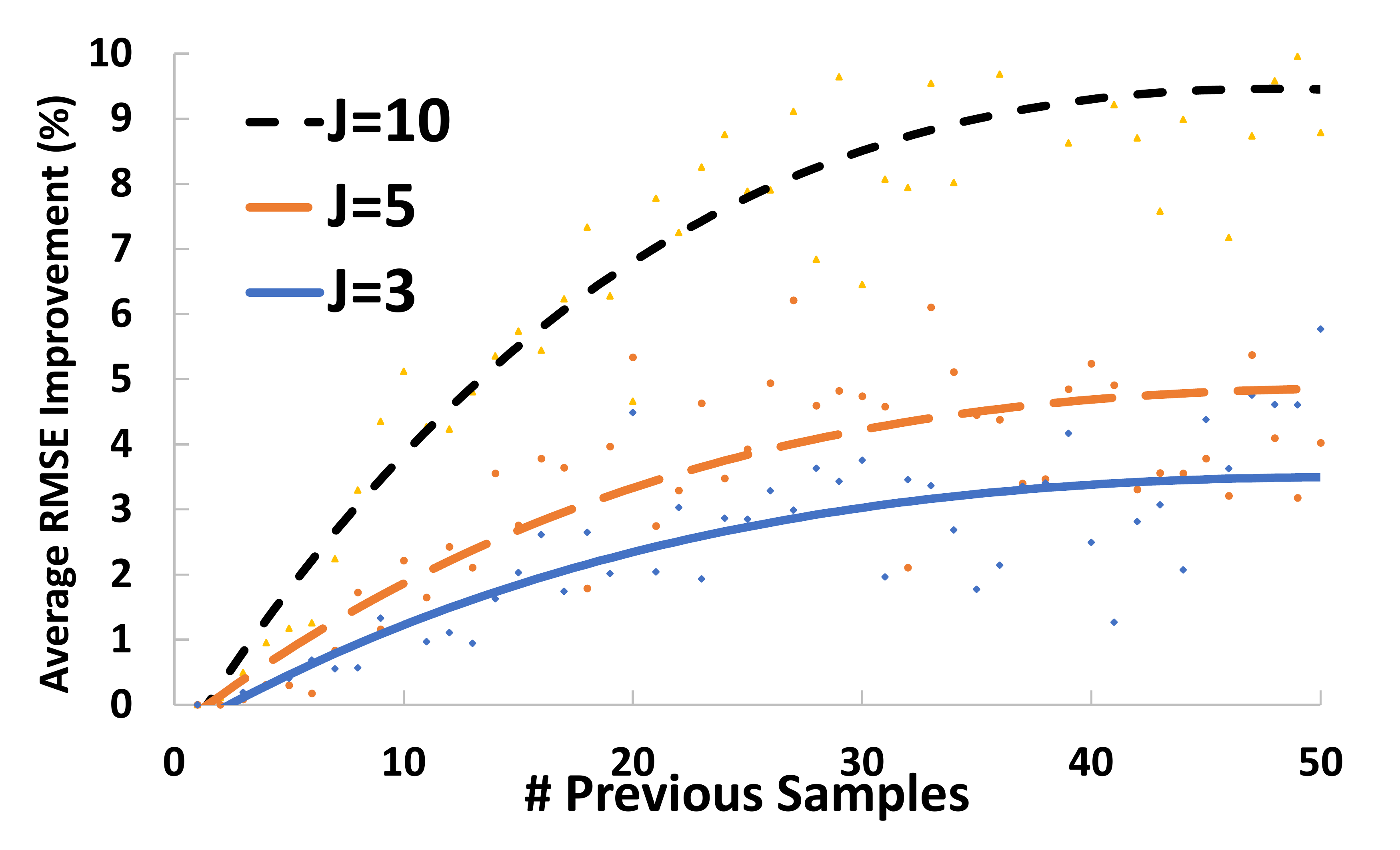}
        \vspace{-1.5mm}
        \caption{Matern kernel}
        \label{fig:Matern_kernel}
    \end{subfigure}
    \begin{subfigure}[b]{0.24\textwidth}
        \includegraphics[scale=0.08]{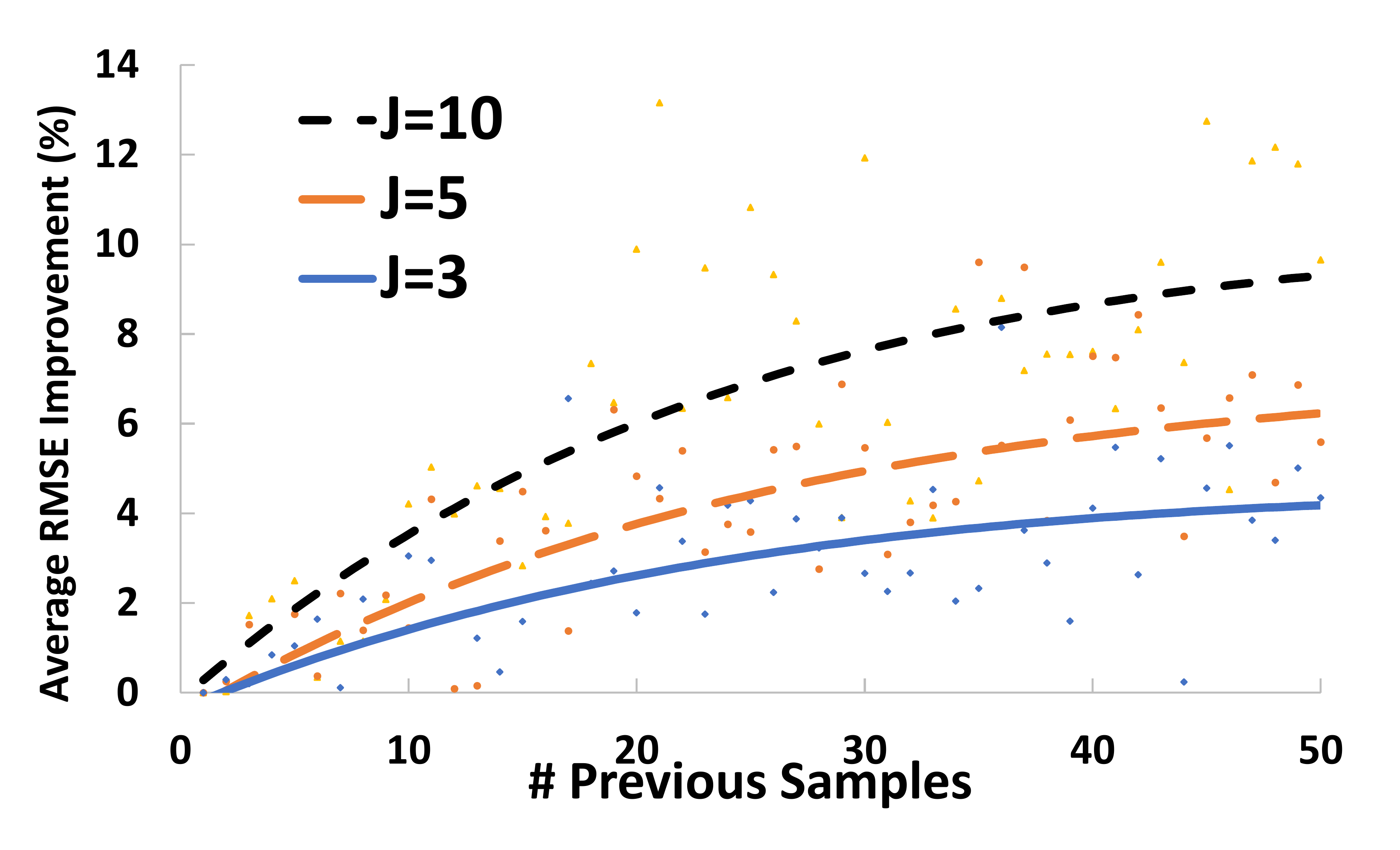}
        \vspace{-1.5mm}
        \caption{Rational quadratic}
        \label{fig:RQ_kernel}
    \end{subfigure}
    \begin{subfigure}[b]{0.24\textwidth}
        \includegraphics[scale=0.08]{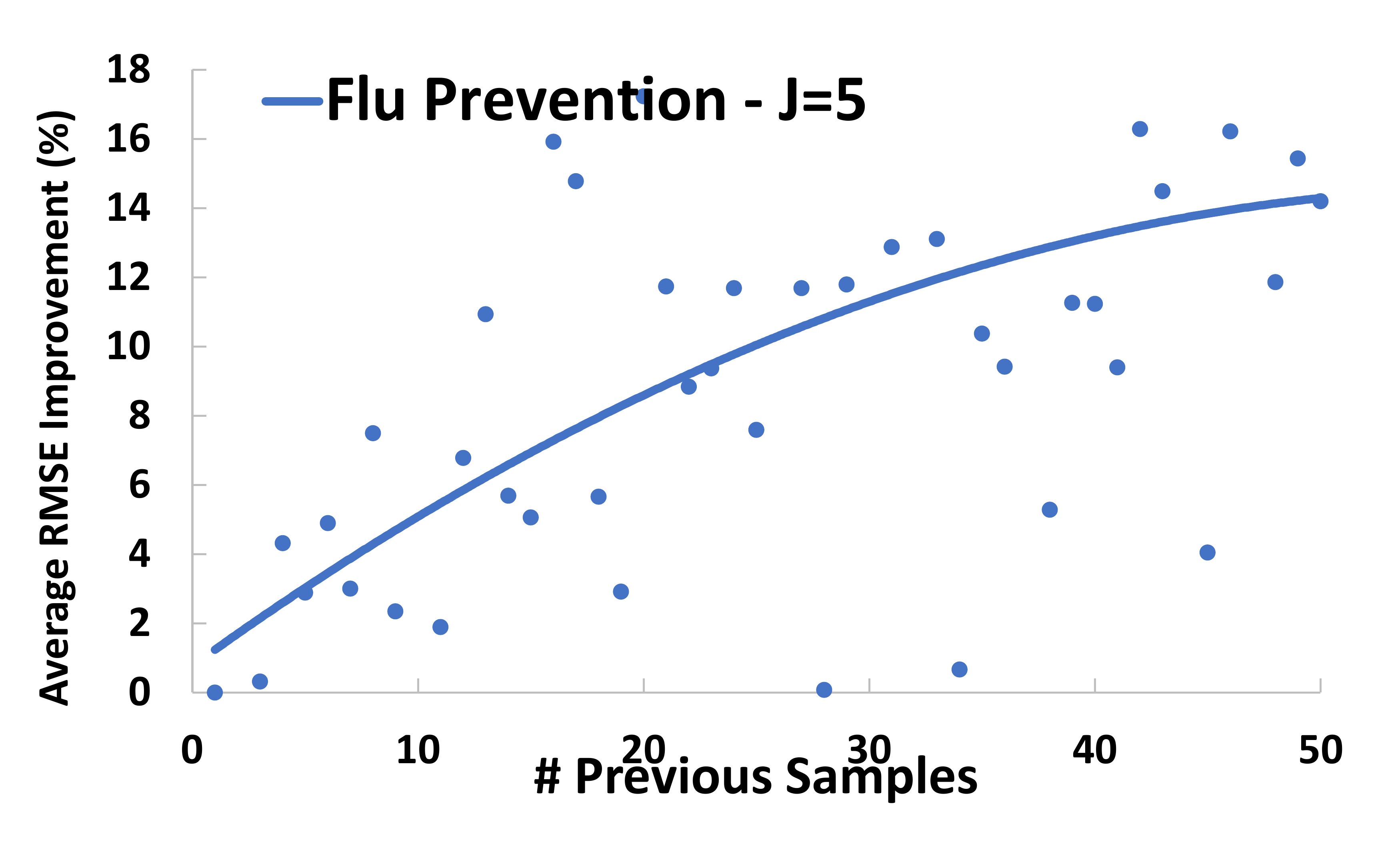}
        \vspace{-1.5mm}
        \caption{Flu domain}
        \label{fig:flu_kernel}
    \end{subfigure}
    \vspace{-3mm}
    \caption{Average improvement for different kernels (with trend line) using decomposed GP regression and GP regression, in RMSE}
	\label{fig:gpr_exp}
\end{figure}

In this section, we run several experiments to compare decomposed GP regression (Algorithm \ref{alg:decomposed_gpr}), D-GPUCB (Algorithm \ref{alg:decomposed_gpucb}), and generalized D-GPUCB (Algorithm \ref{alg:generalized_decomposed_gpucb})
\footnote{Our implementation can be found on Github: https://github.com/guaguakai/DGPUCB}.
We also test on both discrete sample space and continuous sample space.
All of our examples show a promising convergence rate and also improvement against the GP-UCB algorithm, again demonstrating that more detailed modeling reduces the predictive uncertainty.
We use the following three common types of stationary kernel \cite{rasmussen2004gaussian}:

\begin{itemize}
\item The Square Exponential kernel is
$k(\boldsymbol{x}, \boldsymbol{x}') = $
$\exp(-(2l^2)^{-1} \left\Vert \boldsymbol{x} - \boldsymbol{x}' \right\Vert^2)$, $l$ is a length-scale hyper parameter.
\item The Mat\'ern kernel is given by $k(\boldsymbol{x}, \boldsymbol{x}') = (2^{1-\nu} / \Gamma(\nu)) r^\nu B_\nu(r), r = (\sqrt[]{2 \nu} / l) \left\Vert \boldsymbol{x} - \boldsymbol{x}' \right\Vert$, where $\nu$ controls the smoothness of sample functions and $B_\nu$ is a modified Bessel function.
\item The Rational Quadratic kernel is $k(\boldsymbol{x}, \boldsymbol{x}') = (1 + \left\Vert \boldsymbol{x} - \boldsymbol{x}' \right\Vert^2 / (2 \alpha l^2))^{- \alpha}$.
It can be seen as a scale mixture of square exponential kernels with different length-scales.
\end{itemize}

\subsection{Decomposed Gaussian Process Regression}
For the decomposed Gaussian process regression, we perform both GP regression and decomposed GP regression (Algorithm \ref{alg:decomposed_gpr}) on an unknown function that can be linearly decomposed into subfunctions.
Then we compare the root-mean-squared error (RMSE) between the true underlying function and the estimate given by either GP regression or decomposed GP regression.
We run the experiments on synthetic and flu domain.

\subsubsection{Synthetic Data}
For each kernel category, we first draw $J$ kernels with random hyper-parameters.
The sample space $\mathcal{X} = [0,1]$ is uniformly discretized into 1000 points.
We then generate a random sampled function $f_j$ from each corresponding kernel $k_j$ as the target function, combined with the simplest linear decomposition (Definition \ref{def:decomposition}) with $g_j(\boldsymbol{x}) \equiv 1 \forall j$.
For each setting and each $T \leq 50$, we randomly draw $T$ samples as the previous samples and perform GP regression on the composed kernel and decomposed GP regression on the kernels.
We record the average improvement in terms of RMSE against the underlying target function over $100$ independent runs and random subfunctions $f_j$ for each setting.
The results are shown in Figure \ref{fig:RBF_kernel}, \ref{fig:Matern_kernel}, \ref{fig:RQ_kernel}, where the x-axis represents the number of prior samples and y-axis shows the average improvement on RMSE in percentage.

\subsubsection{Flu Prevention}
Based on prior work \cite{del2013mathematical} on disease spread and historic data collected from the CDC \cite{cdc}, we model the flu spread with an age-stratified SIR compartment model, which is a model commonly used to simulate the disease interactions between different age groups.
We run experiments with a square exponential kernels and the result is illustrated in Figure \ref{fig:flu_kernel}.
This domain involves linear decomposition into age groups. 
Additional detail about flu and our model will be provided in section \ref{sec:fluPrevSec}.

Empirically, our method reduces the RMSE in the predictions by 10-15\% compared to standard GP regression (without decomposed feedback). This trend holds across kernels, and includes both synthetic data and the flu domain (which uses a real dataset).
Such an improvement in predictive accuracy is significant in many real-world domains.
For instance, CDC-reported 95\% confidence intervals for vaccination-averted flu illnesses for 2015 \cite{cdc_averted} range from 3.5M-7M and averted medical visits from 1.7M-3.5M.
Reducing average error by 10\% corresponds to estimates which are tighter by hundreds of thousands of patients, a significant amount in policy terms.

In general, the benefit of using decomposed feedback in the regression increases with the number of composed functions $J$, i.e., the benefit from our approach grows with the number of subgroups.
This is a direct result of Theorem \ref{thm:reduced_variance}, where more decomposed feedback tends to have a larger Jensen gap.
These results confirm our theoretical analysis by showing that incorporating decomposed feedback results in more accurate estimation of the unknown function.


\subsection{Comparison between GP-UCB and D-GPUCB}
\begin{figure}[t]
    \centering
    \begin{subfigure}[b]{0.32\textwidth}
        \includegraphics[width=4.2cm,height=2.1cm]{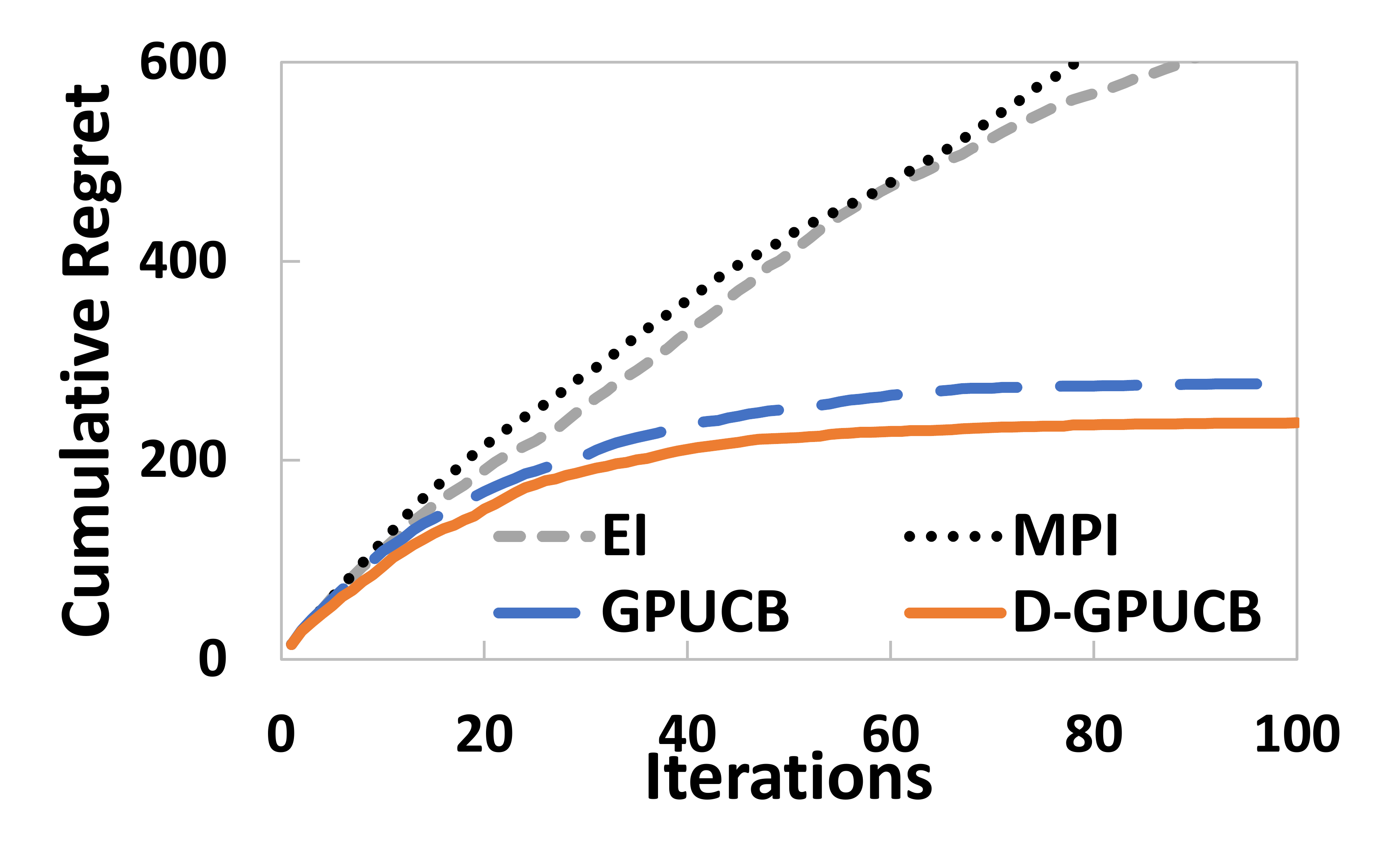}
        \vspace{-7mm}
        \caption{Synthetic data}
        \label{fig:synthetic_cumulative_exp}
    \end{subfigure}
    \begin{subfigure}[b]{0.32\textwidth}
        \includegraphics[width=4.2cm,height=2.1cm]{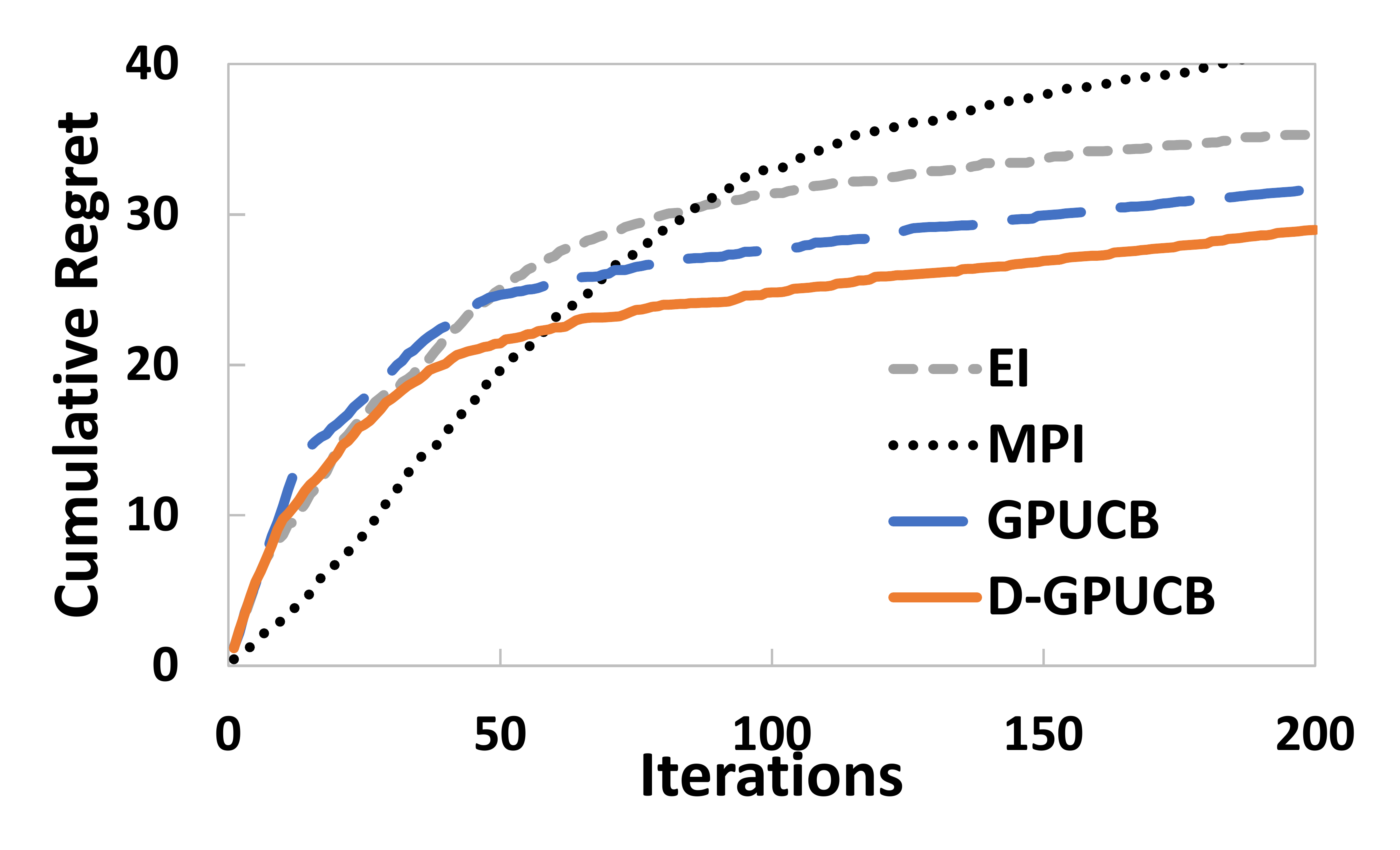}
        \vspace{-7mm}
        \caption{Flu prevention}
        \label{fig:flu_cumulative_exp}
    \end{subfigure}
    \begin{subfigure}[b]{0.32\textwidth}
        \includegraphics[width=4.2cm,height=2.1cm]{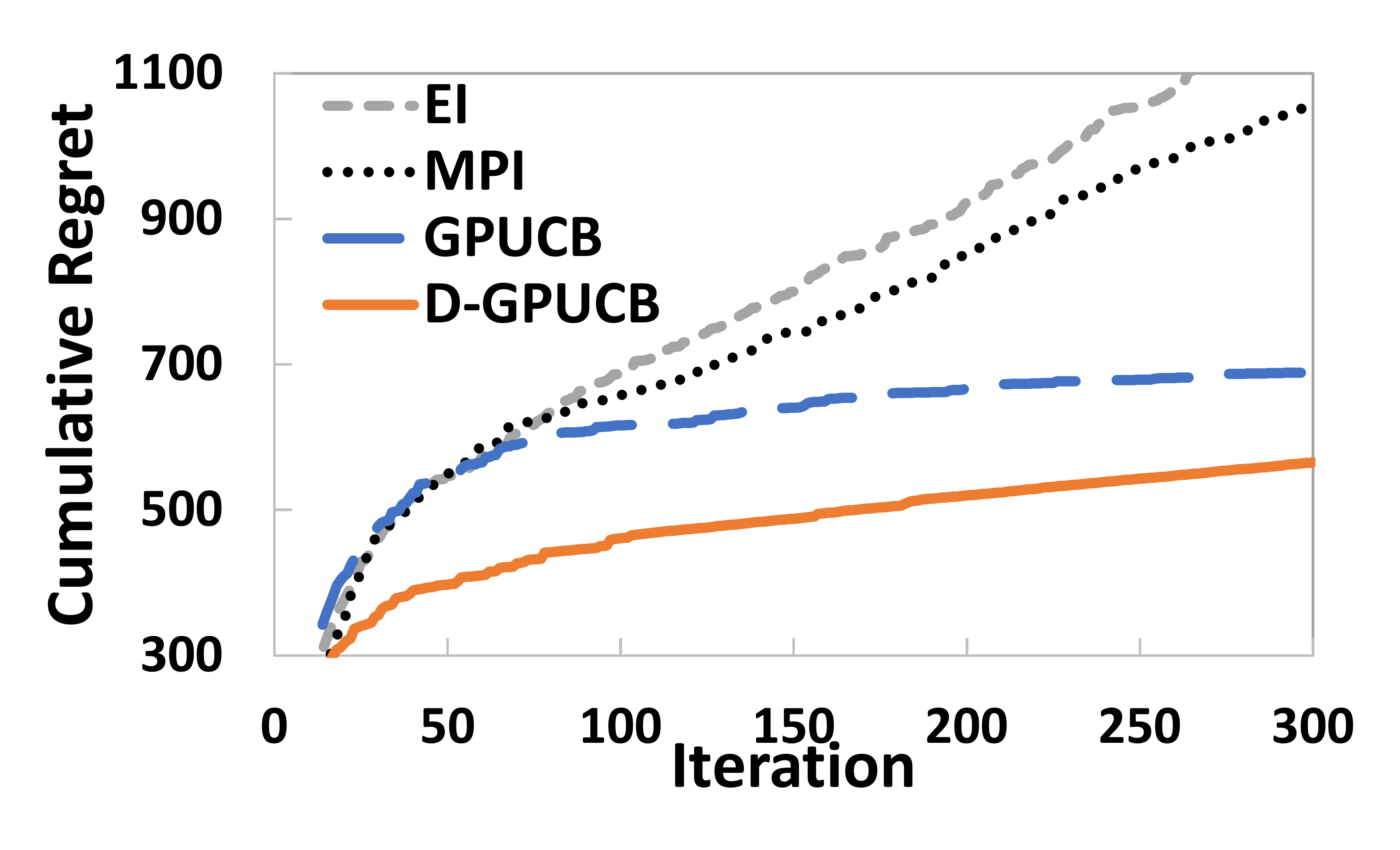}
        \vspace{-7mm}
        \caption{Perceived temperature}
        \label{fig:weather_cumulative_exp}
    \end{subfigure}
    \vspace{-3mm}
    \caption{Comparison of cumulative regret: D-GPUCB, GP-UCB, and various heuristics on synthetic (a) and real data (b, c)}
    \label{fig:exp:decomposed_gpucb_cumulative}
\end{figure}
\begin{figure}[t]
    \centering
    \begin{subfigure}[b]{0.32\textwidth}
        \includegraphics[width=4.2cm,height=2.1cm]{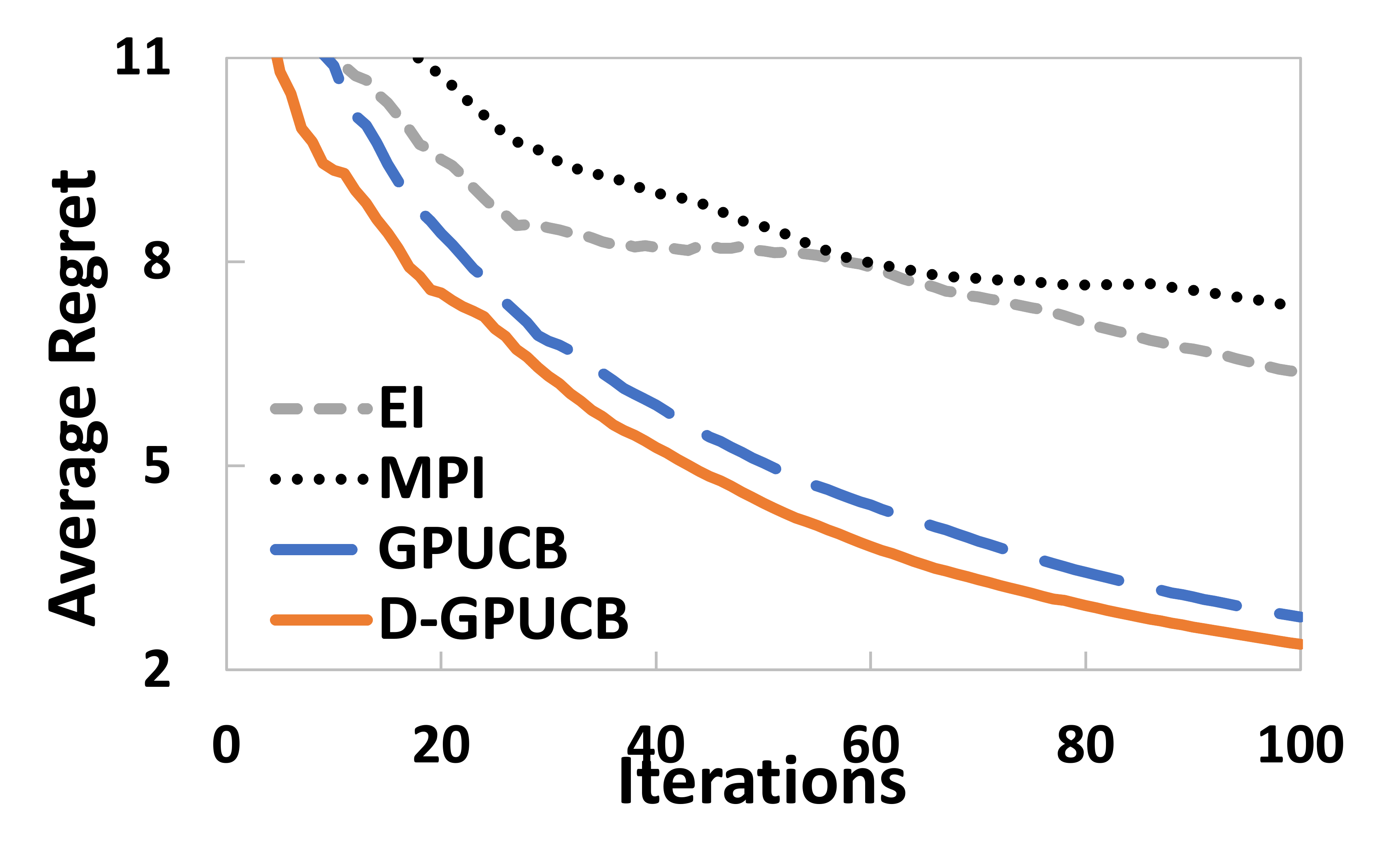}
        \vspace{-7mm}
        \caption{Synthetic data}
        \label{fig:synthetic_exp}
    \end{subfigure}
    \begin{subfigure}[b]{0.32\textwidth}
        \includegraphics[width=4.2cm,height=2.1cm]{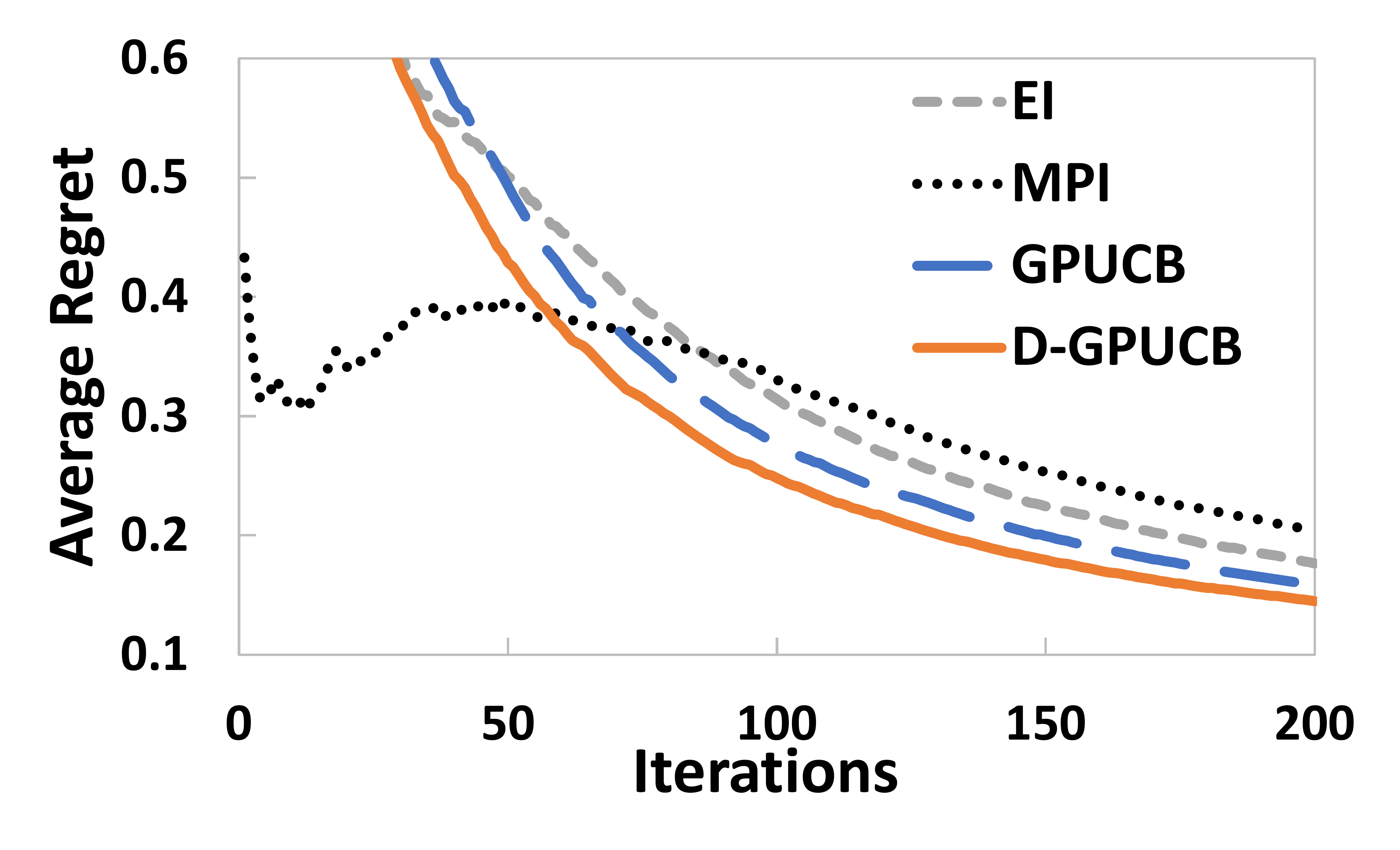}
        \vspace{-7mm}
        \caption{Flu prevention}
        \label{fig:flu_exp}
    \end{subfigure}
    \begin{subfigure}[b]{0.32\textwidth}
        \includegraphics[width=4.2cm,height=2.1cm]{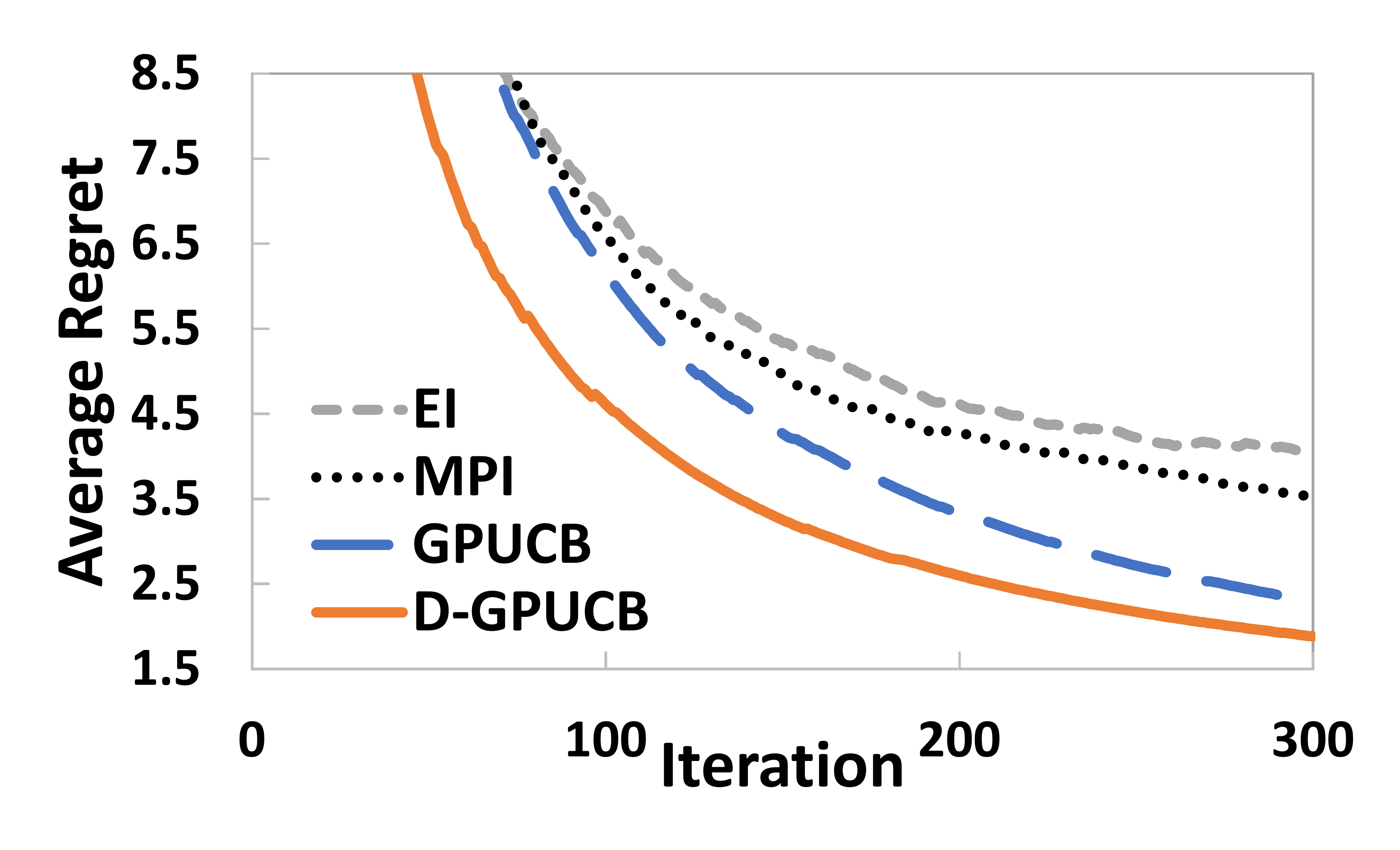}
        \vspace{-7mm}
        \caption{Perceived temperature}
        \label{fig:weather_exp}
    \end{subfigure}
    \vspace{-3mm}
    \caption{Comparison of average regret: D-GPUCB, GP-UCB, and various heuristics on synthetic (a) and real data (b, c)}
	\label{fig:exp_decomposed_gpucb}
\end{figure}

We now move to the online setting to test whether greater predictive accuracy results in improved decision making. We compare our D-GPUCB algorithm and generalized D-GPUCB with GP-UCB, as well as common heuristics such as Expected Improvement (EI) \cite{movckus1975bayesian} and Most Probable Improvement (MPI) \cite{kushner1964new}.
We follow the recommendation in \cite{srinivas2009gaussian} to scale down $\beta_t$ by a factor $5$ for both GP-UCB and D-GPUCB algorithm.
For all the experiments, we run 30 trials on all algorithms to find the average regret.

\subsubsection{Synthetic Data (Linear Decomposition with Discrete Sample Space):}
For synthetic data, we randomly draw $J=10$ square exponential kernels with different hyper-parameters and then sample random functions from these kernels to compose the entire target function.
The sample noise variance is set to be $10^{-4}$.
We run each algorithm for 100 iterations with $\delta = 0.05$ (different kernels and target functions each trial), where the cumulative regrets are shown in Figure \ref{fig:synthetic_cumulative_exp}, and average regret in Figure \ref{fig:synthetic_exp}.

\subsubsection{Flu Prevention (Linear Decomposition with Continuous Sample Space):}\label{sec:fluPrevSec}
We consider an age-stratified flu SIR model \cite{del2013mathematical} as our target function.
The population is stratified into several age groups: young (0-19), adult (20-49), middle aged (50-64), senior (65-69), elder (70+).
The SIR model allows the contact matrix and susceptibility of each age group to vary.
We use the contact matrix in \cite{del2013mathematical}, where the authors learn a contact matrix from a population of over 1.6 million simulated individuals from Portland, Oregon.
Our model is built using the entire US population in 2012, given by the Current Population Survey, US Department of Commerce and with the initial infected population used in \cite{del2013mathematical}.
An individual's susceptibility to disease in different age groups is given by \cite{del2013mathematical} and is calibrated from the data collected by the Center for Disease Control and Prevention \cite{cdc}.
We consider this SIR model to be our target function, and we perform our D-GPUCB and other algorithms to find the optimum.

The controlled input here is the vaccination rate $\boldsymbol{x} \in [0,1]^5$ with respect to each age group.
Given an input $\boldsymbol{x}$, the SIR model returns the average sick days per person $f(\boldsymbol{x})$ within one year.
The model also returns the contribution to the average sick days from each age group $j$, which we denote as $f_j(\boldsymbol{x})$.
Therefore we have $f(\boldsymbol{x}) = \sum\nolimits_{j=1}^5 f_j(\boldsymbol{x})$, a linear decomposition.
The goal is to find the optimal vaccination policy which minimizes the average sick days subject to budget constraints.
In order to get an estimate on the unknown kernel functions, we randomly draw 1000 samples to compute the empirical covariance function by turning hyper-parameters.
We run all algorithms and compare their cumulative regret in Figure \ref{fig:flu_cumulative_exp} and average regret in Figure \ref{fig:flu_exp}.

\subsubsection{Perceived Temperature (General Decomposition with Discrete Sample Space):}
The perceived temperature is a combination of actual temperature, humidity, and wind speed.
The perceived temperature formula has been widely studied and is usually identified using multivariate regression on human preference testing data \cite{kleinbaum2013applied}.
Several models \cite{lee1980seventy,andrade2011perception} have also been proposed with nonlinear terms, and multivariate polynomial models are common.
We use the heat index \cite{steadman1979assessment} and wind chill factor \cite{osczevski2005new} in our perceived temperature model.
When the actual temperature is high, higher humidity reduces the body's ability to cool itself, resulting a higher perceived temperature; when the actual temperature is low, the air motion accelerates the rate of heat transfer from a human body to the surrounding atmosphere, leading to a lower perceived temperature.
So the perceived temperature is a multivariate polynomial of humidity (wind speed) and actual temperature when the actual temperature is relatively high (low), which are all nonlinear function compositions.
We use weather data collected from 2906 sensors in the United States provided by OpenWeatherMap \cite{openweathermap} on 14, Mar 2017.
In each independent test, we follow the experimental setting in \cite{srinivas2009gaussian} to randomly select one-third of the sensors to fit the kernel function, then perform generalized D-GPUCB and all the other algorithms on the remaining sensors: where given an input location $\boldsymbol{x} \in \mathcal{X}$, we can access the actual temperature $f_1(\boldsymbol{x})$, humidity $f_2(\boldsymbol{x})$, and wind speed $f_3(\boldsymbol{x})$.
The goal is to find the location with highest perceived temperature in the remaining sensors.
The result is averaged over 30 different tests and is also shown in Figure \ref{fig:weather_cumulative_exp} and Figure \ref{fig:weather_exp}.

\subsection{Discussion}
In the bandit setting with decomposed feedback, Figure \ref{fig:exp:decomposed_gpucb_cumulative} shows a $10\%-20\%$ improvement in cumulative regret for both synthetic (Figure \ref{fig:synthetic_cumulative_exp}) and real data (Figure \ref{fig:flu_cumulative_exp}, \ref{fig:weather_cumulative_exp}).
As in the regression setting, such improvements are highly significant in policy terms; a 10\% reduction in sickness due to flu corresponds to hundreds of thousands of infections averted per year.

The benefit to incorporating decomposed feedback is particularly large in the general decomposition case (Figure \ref{fig:weather_cumulative_exp}), where a single GP is a poor fit to the non-linearly composed function.
The non-linearity of the composition leads to a non-Gaussian composed function, in which case GP-UCB cannot guarantee any theoretical regret bound. Confirming this theoretical result, we observe a large gap between D-GPUCB, a method with theoretical guarantees, and GP-UCB, without guarantees.
This showcases the benefit of decomposed feedback, especially in a non-linear, complicated system.
Figure \ref{fig:exp_decomposed_gpucb} shows the average regret of each algorithm (as opposed to the cumulative regret).
Our algorithm's average regret tends to zero.
This empirically confirms the no-regret guarantee for D-GPUCB in both the linear and general decomposition settings.


\section{Conclusions}
In this paper, we proposed a decomposed approach that explicitly allows Bayesian approaches to exploit the information in subgroup dynamics, which captures the heterogeneity of subgroups through decomposed feedback.
These approaches also preserve the ability to optimize over the underlying complex systems.
This pushes the boundary of current work on the trade-off between using more precise simulations and Bayesian approaches with theoretical guarantee.


In the regression setting, we propose a decomposed GP regression (Algorithm \ref{alg:decomposed_gpr}) and prove that incorporating decomposed feedback improves predictive accuracy (Theorem \ref{thm:reduced_variance}).
In the online learning setting, we introduce the D-GPUCB algorithms and generalized D-GPUCB (Algorithm \ref{alg:decomposed_gpucb} and Algorithm \ref{alg:generalized_decomposed_gpucb}) with corresponding no-regret guarantees.
Moreover, generalized D-GPUCB extends the applicability of Bayesian approaches to non-linear composition of Gaussian processes, which cannot be dealt by the previous methods.
We conduct experiments in both real and synthetic domains to investigate the performance of decomposed GP regression, D-GPUCB, and generalized D-GPUCB.
All of the results show significant improvement against GP-UCB and other methods that do not consider decomposed feedback, demonstrating the benefit that decision makers can realize by exploiting such information. 


%
%
%
\bibliographystyle{splncs04}
\bibliography{reference}
\newpage
\section{Appendix}
\entireVariance*
\begin{proof}[Proof of Proposition \ref{prop:entire_variance}]

According to Equation \eqref{eqn:gpr_covar}, the posterior covariance $k_{T, \text{entire}}(\boldsymbol{x}, \boldsymbol{x}')$ in the standard GP regression can be written as:
\begin{align}\label{eqn:entire_posterior_covariance}
k_{T, \text{entire}}(\boldsymbol{x}, \boldsymbol{x}') &= k(\boldsymbol{x}, \boldsymbol{x}') - \boldsymbol{k}_T(\boldsymbol{x})^\top {\boldsymbol{K}_T}^{-1} \boldsymbol{k}_T(\boldsymbol{x})
\end{align}

By the decomposition assumption (Equation \eqref{eqn:decomposition}), we have $k(\boldsymbol{x}, \boldsymbol{x}') = \sum\limits_{j=1}^J g_j(\boldsymbol{x}) k_j(\boldsymbol{x}, \boldsymbol{x}') g_j(\boldsymbol{x}')$.
Moreover,
\begin{align}
&\boldsymbol{k}_T(\boldsymbol{x}) = [k(\boldsymbol{x}_1, \boldsymbol{x}), ..., k(\boldsymbol{x}_T, \boldsymbol{x})]^\top \nonumber \\
&= \sum\limits_{j=1}^J [g_j(\boldsymbol{x}_1) k_j(\boldsymbol{x}_1, \boldsymbol{x}) g_j(\boldsymbol{x}), ..., g_j(\boldsymbol{x}_T) k_j(\boldsymbol{x}_T, \boldsymbol{x}) g_j(\boldsymbol{x})]^\top \nonumber \\
&= \sum\limits_{j=1}^J \boldsymbol{D}_j \boldsymbol{k}_{j,T}(\boldsymbol{x}) g_j(\boldsymbol{x}) \label{eqn:posterior_covariance_vector}
\end{align}
where $\boldsymbol{k}_{j,T}(\boldsymbol{x}) = [k_j(\boldsymbol{x}_1, \boldsymbol{x}), ..., k_j(\boldsymbol{x}_T, \boldsymbol{x})]^\top$.

The variance function $\sigma_T^2(\boldsymbol{x})$ is just the value of covariance function with $\boldsymbol{x}'= \boldsymbol{x}$.
Therefore, combining Equation \eqref{eqn:entire_posterior_covariance} and \eqref{eqn:posterior_covariance_vector}, the variance can be written as:
\begin{align}
&\sigma_{T, \text{entire}}^2(\boldsymbol{x}) = k_{T, \text{entire}}(\boldsymbol{x}, \boldsymbol{x}) \nonumber \\
&= k(\boldsymbol{x}, \boldsymbol{x}) - \boldsymbol{k}_T(\boldsymbol{x})^\top \boldsymbol{K}_T^{-1} \boldsymbol{k}_T(\boldsymbol{x}) \nonumber \\ 
&= k(\boldsymbol{x}, \boldsymbol{x}) - \sum\limits_{i,j} g_i(\boldsymbol{x}) \boldsymbol{k}_{i,T}(\boldsymbol{x})^\top \boldsymbol{D}_i^\top \boldsymbol{K}_T^{-1} \boldsymbol{D}_{j} \boldsymbol{k}_{j,T}(\boldsymbol{x}) g_{j}(\boldsymbol{x}) \nonumber \\
&= k(\boldsymbol{x}, \boldsymbol{x}) - \sum\limits_{i,j} \boldsymbol{z}_i^\top \boldsymbol{K}_T^{-1} \boldsymbol{z}_j \label{eqn:kernel_matrix_unexpanded} \\
&= k(\boldsymbol{x}, \boldsymbol{x}) - \sum\limits_{i,j} \boldsymbol{z}_i^\top (\sum\limits_{l} \boldsymbol{D}_l \boldsymbol{K}_{l,T} \boldsymbol{D}_l)^{-1} \boldsymbol{z}_j \label{eqn:kernel_matrix_expanded}
\end{align}
with $\boldsymbol{z}_i = \boldsymbol{D}_i \boldsymbol{k}_{j,T} g_j(\boldsymbol{x}) \in \R^n$ and equation \eqref{eqn:kernel_matrix_unexpanded} to \eqref{eqn:kernel_matrix_expanded} is coming from:
\begin{align}
\boldsymbol{K}_T &= [k(\boldsymbol{x},\boldsymbol{x}')]_{\boldsymbol{x},\boldsymbol{x}' \in A_T} + \text{diag}([\sigma^2(\boldsymbol{x}_t)]_{t \in [T]}) \label{eqn:posterior_covariance_matrix_1} \\
&= \sum\limits_{j=1}^J [ g_j(\boldsymbol{x}) k_j(\boldsymbol{x},\boldsymbol{x}') g_j(\boldsymbol{x}')]_{\boldsymbol{x},\boldsymbol{x}' \in A_T} \nonumber \\ 
& \qquad \qquad + \text{diag}([g_j^2(\boldsymbol{x}_t) \sigma^2_j(\boldsymbol{x}_t)]_{t \in [T]}) \label{eqn:posterior_covariance_matrix_2} \\
&= \sum\limits_{j} \boldsymbol{D}_j ([k_j(\boldsymbol{x},\boldsymbol{x}') ]_{\boldsymbol{x},\boldsymbol{x}' \in A_T} + \text{diag}([\sigma^2_j(\boldsymbol{x}_t)]_{t \in [T]})) \boldsymbol{D}_j \nonumber \\
&= \sum\limits_{j} \boldsymbol{D}_j \boldsymbol{K}_{j,T} \boldsymbol{D}_j \nonumber
\end{align}
where the first kernel term from Equation \eqref{eqn:posterior_covariance_matrix_1} to \eqref{eqn:posterior_covariance_matrix_2} is derived by from definition.
And in the latter term, from the decomposition assumption \eqref{eqn:decomposition}, the noise variance $\sigma^2(\boldsymbol{x})$ of the target function $f$ at point $\boldsymbol{x}$ is the cumulative variance of the noise variance $\sigma_j^2(\boldsymbol{x})$ of each individual function $f_j$, i.e.
\begin{equation*}
\sigma^2(\boldsymbol{x}) = \sum\limits_{j=1}^J g_j^2(\boldsymbol{x}) \sigma^2_j(\boldsymbol{x}) ~\forall \boldsymbol{x} \in \mathcal{X}, j \in [J]
\end{equation*}
which explains the derivation from Equation \eqref{eqn:posterior_covariance_matrix_1} to \eqref{eqn:posterior_covariance_matrix_2}.
\qed
\end{proof}

\decomposedVariance*
\begin{proof}[Proof of Proposition \ref{prop:decomposed_variance}]

In Algorithm \ref{alg:decomposed_gpr}, it runs GP regression to each function $f_j(\boldsymbol{x})$ respectively.
We can compute the corresponding posterior covariance function $k_{j,T}$ by:
\begin{equation*}
k_{j,T}(\boldsymbol{x}, \boldsymbol{x}') = k_j(\boldsymbol{x}, \boldsymbol{x}') - \boldsymbol{k}_{j,T}(\boldsymbol{x})^\top \boldsymbol{K}_{j,T}^{-1} \boldsymbol{k}_{j,T}(\boldsymbol{x})
\end{equation*}
By Algorithm \ref{alg:decomposed_gpr}, the synthetic covariance of the target function $f(\boldsymbol{x})$ is:
\begin{align*}
k_{T, \text{decomp}}(\boldsymbol{x}, \boldsymbol{x}') = \sum\limits_{j=1}^J g_j(\boldsymbol{x}) k_{j,T}(\boldsymbol{x}, \boldsymbol{x}') g_j(\boldsymbol{x}')
\end{align*}
\begin{align*}
&\sigma_{T, \text{decomp}}^2(\boldsymbol{x}) = k_{T, \text{decomp}}(\boldsymbol{x}, \boldsymbol{x}) \\
&= \sum\limits_{j=1}^J g_j(\boldsymbol{x}) k_{j,T}(\boldsymbol{x}, \boldsymbol{x}) g_j(\boldsymbol{x}) \\
&= \sum\limits_{j=1}^J g_j(\boldsymbol{x}) k_j(\boldsymbol{x}, \boldsymbol{x}) g_j(\boldsymbol{x}) \\
& \qquad \qquad - \sum\limits_{j=1}^J g_j(\boldsymbol{x}) \boldsymbol{k}_{j,T}(\boldsymbol{x})^\top \boldsymbol{K}_{j,T}^{-1} \boldsymbol{k}_{j,T}(\boldsymbol{x}) g_j(\boldsymbol{x}) \\
&= k(\boldsymbol{x}, \boldsymbol{x}) - \sum_{j} \boldsymbol{z}_j^\top \boldsymbol{D}_j^{-1} \boldsymbol{K}_{j,T}^{-1} \boldsymbol{D}_j^{-1} \boldsymbol{z}_j \\
&= k(\boldsymbol{x}, \boldsymbol{x}) - \sum_{j} \boldsymbol{z}_j^\top (\boldsymbol{D}_j \boldsymbol{K}_{j,T} \boldsymbol{D}_j)^{-1} \boldsymbol{z}_j
\end{align*}
\qed
\end{proof}

\section{Proof of Theorem \ref{thm:reduced_variance}}
\reducedVariance*

\begin{proof}
If we write $\boldsymbol{B}_l = \boldsymbol{D}_l \boldsymbol{K}_{l,T} \boldsymbol{D}_l$, then $\boldsymbol{B}_l$ is positive definite since it is the multiplication of positive definite matrix $\boldsymbol{K}_{l,T}$ and two $\boldsymbol{D}_l$ identical diagonal matrices.
Using Proposition \ref{prop:entire_variance} and \ref{prop:decomposed_variance}, the difference between Equation \eqref{eqn:entire_variance} and \eqref{eqn:decomposed_variance} can be written as:
\begin{align*}
&\sigma_{T, \text{entire}}^2(\boldsymbol{x}) - \sigma_{T, \text{decomp}}^2(\boldsymbol{x}) \\
&=\! \sum_{l} \boldsymbol{z}_l^\top (\boldsymbol{D}_l \boldsymbol{K}_{l,T} \boldsymbol{D}_l)^{-1} \boldsymbol{z}_l \! - \! \sum_{i,j} \boldsymbol{z}_i^\top (\sum_{l} \boldsymbol{D}_l \boldsymbol{K}_{l,T} \boldsymbol{D}_l)^{-1} \boldsymbol{z}_j \\
&= \sum\nolimits_{l} \boldsymbol{z}_l^\top \boldsymbol{B}_l^{-1} \boldsymbol{z}_l - \sum\nolimits_{i,j} \boldsymbol{z}_i^\top (\sum\nolimits_{l} \boldsymbol{B}_l )^{-1} \boldsymbol{z}_j \\
&= \sum\nolimits_{l} \boldsymbol{z}_l^\top \boldsymbol{B}_l^{-1} \boldsymbol{z}_l - J (\frac{\sum\nolimits_{i} \boldsymbol{z}_i}{J})^\top (\frac{\sum_{l} \boldsymbol{B}_l}{J})^{-1} (\frac{\sum\nolimits_{i} \boldsymbol{z}_i}{J}) \\
&= \sum\nolimits_{l} h(\boldsymbol{B}_l, \boldsymbol{z}_l) - J h(\bar{\boldsymbol{B}}, \bar{\boldsymbol{z}}) \geq 0
\end{align*}
where $\bar{\boldsymbol{B}} = \frac{\sum\nolimits_i \boldsymbol{B}_i}{J}$ and $\bar{\boldsymbol{z}} = \frac{\sum\nolimits_i \boldsymbol{z}_i}{J}$ are the average value.
The last inequality comes from Jensen inequality and Lemma \ref{lemma:matrix_fractinoal_function}, which says the matrix-fractional function $h$ is convex.
\qed
\end{proof}

\section{Proof of Theorem \ref{thm:decomposed_GPUCB_bound}}

In order to prove this, we follow the similar techniques as GPUCB \cite{srinivas2009gaussian}, which is illustrated as follows:

\begin{lemma}[Modified version of Lemma 5.1 from \cite{srinivas2009gaussian}]\label{lemma:chernoff_bound_modified}
Given $f(\boldsymbol{x}) = \sum\nolimits_{j=1}^J g_j(\boldsymbol{x}) f_j(\boldsymbol{x})$ (Definition \ref{def:decomposition}), deterministic known functions $g_j$ and unknown $f_j \sim GP(0, k_j(\boldsymbol{x}, \boldsymbol{x}'))$, pick $\delta \in (0,1)$ and set $\beta_t = 2 \log(|\mathcal{X}| \pi_t / \delta)$, where $\sum\nolimits_{t\geq 1 } \pi_t^{-1} = 1, \pi_t > 0 $. Then, the $\mu_{t-1}(\boldsymbol{x}), \sigma_{t-1}(\boldsymbol{x})$ returned by Algorithm \ref{alg:decomposed_gpucb} satisfy:
\begin{align*}
|f(\boldsymbol{x}) - \mu_{t-1}(\boldsymbol{x})| \leq \beta_t^{1/2} \sigma_{t-1}(\boldsymbol{x}) \forall \boldsymbol{x} \in \mathcal{X}, t \geq 1
\end{align*}
with probability $1 - \delta$.
\end{lemma}
\begin{proof}
Fix $t \geq 1$ and $\boldsymbol{x} \in \mathcal{X}$, Conditioned on sampled points $\{\boldsymbol{x}_1,...,\boldsymbol{x}_{t-1}\}$ and sampled values $\{ y_{1,j},..., y_{t-1,j} \} \forall j \in [J]$, the Bayesian property of decomposed GP regression (Algorithm \ref{alg:decomposed_gpr}) implies that the function value at point $\boldsymbol{x}$ forms a Gaussian distribution with mean $\mu_{t-1}(\boldsymbol{x})$ and variance $\sigma^2_{t-1}(\boldsymbol{x})$, i.e. $f(\boldsymbol{x}) \sim N(\mu_{t-1}(\boldsymbol{x}), \sigma^2_{t-1}(\boldsymbol{x}))$.
Now, if $r \sim N(0, 1)$, then
\begin{align}
Pr\{r > c\} &= e^{-c^2/2}(2 \pi)^{-1/2} \int e^{-(r-c)^2 / 2 - c(r-c)} dr \nonumber \\
& \leq e^{-c^2/2} Pr\{r > 0\} = (1/2) e^{-c^2/2} \label{eqn:chernoff_bound}
\end{align}
for $c>0$, since $e^{-c(r-c)} \leq 1$ for $r \geq c$.
Therefore, $Pr\{|f(\boldsymbol{x}) - \mu_{t-1}(\boldsymbol{x})| > \beta_t^{1/2} \sigma_{t-1}(\boldsymbol{x})\} \leq e^{-\beta_t^{1/2}}$, using $r=(f(\boldsymbol{x}) - \mu_{t-1}(\boldsymbol{x}))/\sigma_{t-1}(\boldsymbol{x})$ and $c = \beta_t^{1/2}$. Then apply the union bound to all $\boldsymbol{x} \in \mathcal{X}$:
\begin{align*}
|f(\boldsymbol{x}) - \mu_{t-1}(\boldsymbol{x})| \leq \beta_t^{1/2} \sigma_{t-1}(\boldsymbol{x}) ~\forall \boldsymbol{x} \in \mathcal{X}
\end{align*}
holds with probability $\geq 1 - |\mathcal{X}| e^{-\beta_t^{1/2}}$.
Choosing $|\mathcal{X}| e^{-\beta_t^{1/2}} = \delta/\pi_t$ and using the union bound for $t \in \N$, the statement holds. For example, we can use $\pi_t = \pi^2 t^2 / 6$. 
\qed
\end{proof}

The proof is almost the same as Theorem 5.1 in \cite{srinivas2009gaussian} except the Bayesian property of decomposed Gaussian process, where the Bayesian property of decomposed Gaussian process can be gotten from the Bayesian property of each individual function $f_j$ and the linear combination of Gaussian distributions is still a Gaussian distribution, which implies the posterior belief after performing decomposed GP regression at a given point $\boldsymbol{x}$ still form a Gaussian distribution with composed mean and variance.

\begin{lemma}[Modified version of Lemma 5.2 from \cite{srinivas2009gaussian}]\label{lemma:regret_variance_bound}
Fix $t \geq 1$.
If $|f(\boldsymbol{x}) - \mu_{t-1}(\boldsymbol{x}) | \leq \beta_t^{1/2} \sigma_{t-1}(\boldsymbol{x})$ for all $\boldsymbol{x} \in \mathcal{X}$, then the regret $r_t$ is bounded by $2 \beta_t^{1/2} \sigma_{t-1}(\boldsymbol{x}_t)$, where $\boldsymbol{x}_t$ is the $t$-th choice of Algorithm \ref{alg:decomposed_gpucb}.
\end{lemma}
\begin{proof}
By definition of $\boldsymbol{x}_t$: $\mu_{t-1}(\boldsymbol{x}_t) + \beta_t^{1/2} \sigma_{t-1}(\boldsymbol{x}_t) \geq \mu_{t-1}(\boldsymbol{x}^*) + \beta_t^{1/2} \sigma_{t-1}(\boldsymbol{x}^*) \geq f(\boldsymbol{x}^*)$.
Therefore,
\begin{align*}
r_t &= f(\boldsymbol{x}^*) - f(\boldsymbol{x}_t) \leq \beta_t^{1/2} \sigma_{t-1}(\boldsymbol{x}_t) + \mu_{t-1}(\boldsymbol{x}_t) - f(\boldsymbol{x}_t) \\
&\leq 2 \beta_t^{1/2} \sigma_{t-1}(\boldsymbol{x}_t)
\end{align*}
\qed
\end{proof}

\begin{lemma}[Modified version of Lemma 5.3 from \cite{srinivas2009gaussian}]\label{lemma:information_gain_modified}
The information gain for the points selected can be expressed in terms of the predictive variances.
If $\boldsymbol{f_{j,T}} = \{ f_j(\boldsymbol{x}_t) \}_{t \in [T]} \in \R^T$ and $\boldsymbol{y_{j,T}} = \{y_{j,t} \}_{t \in [T]} \in \R^T$:
\begin{align*}
I(\boldsymbol{y_{j,T}}: \boldsymbol{f_{j,T}}) = \frac{1}{2} \sum\nolimits_{t=1}^T \log(1 + \sigma_j^{-2} \sigma_{j,t-1}^2(\boldsymbol{x}_t))
\end{align*}
where $f_j(\boldsymbol{x}_t), y_{j,t}, \sigma_{j,t-1}^2$ follow the definition and derivation in Algorithm \ref{alg:decomposed_gpucb}.
\end{lemma}
\begin{proof}
Directly follow by replacing all the $f, y, \sigma$ by $f_j, y_j, \sigma_j$ in the proof of Theorem 5.3 from \cite{srinivas2009gaussian}. \qed
\end{proof}

\decomposedGPUCBbound*
\begin{proof}
By choosing an unified individual noise variance $\sigma_j^2 = \sigma^2$, according to Lemma \ref{lemma:information_gain_modified}, we can take advantage of the individual information gain of each $f_j(\boldsymbol{x})$, which is 
\begin{align*}
I_j(y_{j,T}; f_j) &= \frac{1}{2} \sum\limits_{t=1}^T \log (1 + \sigma_j^{-2} \sigma_{j,t-1}^2(\boldsymbol{x}_t) ) \\
&= \frac{1}{2} \sum\limits_{t=1}^T \log (1 + \sigma^{-2} \sigma_{j,t-1}^2(\boldsymbol{x}_t) ) \\
\gamma_{j,T} &= \max I_j(y_{j,T}; f_j)
\end{align*}
Besides, we can also bound the total regret by the individual information gains as following:
\begin{align}
\sum\limits_{j=1}^J B_j^2 I_j(y_{j,T}; f_j) &= \frac{1}{2} \sum\limits_{j=1}^J B_j^2 \sum\limits_{t=1}^T \log(1 + \sigma^{-2} \sigma_{j,t-1}^2 (\boldsymbol{x}_t) ) \nonumber \\
&\geq \frac{1}{2} \sum\limits_{j=1}^J B_j^2 \sum\limits_{t=1}^T C_2^{-1} \sigma^{-2} \sigma_{j,t-1}^2 (\boldsymbol{x}_t) \nonumber \\
&\geq \frac{1}{2} C_2^{-1} \sigma^{-2} \sum\limits_{j=1}^J \sum\limits_{t=1}^T g_j^2(\boldsymbol{x}_t) \sigma_{j,t-1}^2 (\boldsymbol{x}_t) \label{eqn:gj_sigmaj} \\
&\geq \frac{1}{2} C_2^{-1} \sigma^{-2} \sum\limits_{t=1}^T \frac{r_t^2}{4 \beta_t} \label{eqn:regret_beta} \\
&\geq \frac{C_2^{-1} \sigma^{-2}}{8 \beta_T} \sum\limits_{t=1}^T r_t^2 \nonumber
\end{align}
where $C_2 = \sigma^{-2} / \log(1 + \sigma^{-2}) \geq 1$, $s^2 \leq C_2 \log (1 + s^2)$ for $s \in [0, \sigma^{-2}]$ and $\sigma^{-2} \sigma^2_{j,t-1}(\boldsymbol{x}_t) \leq \sigma^{-2} k_j(\boldsymbol{x}_t, \boldsymbol{x}_t) \leq \sigma^{-2}$.
The inequality from \eqref{eqn:gj_sigmaj} to \eqref{eqn:regret_beta} follows by Lemma \ref{lemma:chernoff_bound_modified} and Lemma \ref{lemma:regret_variance_bound}.

Let $C_1 = 8 \sigma^2 C_2 = 8 / \log(1 + \sigma^{-2})$.
Applying Cauchy inequality gives us:
\begin{equation*}
C_1 \beta_T T \sum\limits_{j=1}^J B_j^2 I_j(y_{j,T}; f_j) \geq (\sum\limits_{t=1}^T r_t)^2 = R_T^2
\end{equation*}
which implies a similar upper bound
\begin{equation*}
R_T \leq \sqrt{C_1 T \beta_T \sum\limits_{j=1}^J B_j^2 \gamma_{j,T}}
\end{equation*}
\qed
\end{proof}

\section{Proof of Theorem \ref{thm:generalized_decomposed_GPUCB_bound}}

All the proofs in Theorem \ref{thm:decomposed_GPUCB_bound} apply except Lemma \ref{lemma:chernoff_bound_modified}.
Since the decomposition here is non-linear, therefore the composition of outcomes of Gaussian processes is no longer an outcome of Gaussian process, which prohibits us to have a nice Gaussian process property: function value $f(\boldsymbol{x})$ does not form a Gaussian distribution.
Due to the non-linearity, the distribution gets distorted, losing its original form with Gaussian distribution.
Fortunately, if the partial derivatives of function $g: \R^J \rightarrow \R$ (Definition \ref{def:general_decomposition}) are bounded, then we can still perform a similar estimation and bound the distribution by a larger Gaussian distribution, which enables us to have a similar result.

\begin{lemma}[General Version with Definition \ref{def:general_decomposition} and Algorithm \ref{alg:generalized_decomposed_gpucb}]\label{lemma:lemma:chernoff_bound_modified_generalized}
Given $f(\boldsymbol{x}) = g(f_1(\boldsymbol{x}),...,f_J(\boldsymbol{x}))$ (Definition \ref{def:general_decomposition}), deterministic known functions $g$ and unknown $f_j \sim GP(0, k_j(\boldsymbol{x}, \boldsymbol{x}'))$, pick $\delta \in (0,1)$ and set $\beta_t = 2 \log(|\mathcal{X}| J \pi_t / \delta)$, where $\sum\nolimits_{t\geq 1 } \pi_t^{-1} = 1, \pi_t > 0 $.
Further assume the function $g$ has bounded partial derivatives $B_j = \max\nolimits_{\boldsymbol{x} \in \mathcal{X}} |\nabla_j g(\boldsymbol{x}) | ~\forall j \in [J]$.
Then, the $\mu_{t-1}(\boldsymbol{x}), \sigma_{t-1}(\boldsymbol{x})$ returned by Algorithm \ref{alg:generalized_decomposed_gpucb} satisfy:
\begin{align*}
|f(\boldsymbol{x}) - \mu_{t-1}(\boldsymbol{x})| \leq \beta_t^{1/2} \sigma_{t-1}(\boldsymbol{x}) \forall \boldsymbol{x} \in \mathcal{X}, t \geq 1
\end{align*}
with probability $1 - \delta$.
\end{lemma}
\begin{proof}
The main issue here is the posterior distribution of $f(\boldsymbol{x})$ is not a Gaussian distribution.
But fortunately, the posterior distribution of each $f_j(\boldsymbol{x})$ is still a Gaussian distribution with mean $\mu_{j,t-1}(\boldsymbol{x})$ and variance $\sigma^2_{j,t-1}(\boldsymbol{x})$ for any given $\boldsymbol{x} \in \mathcal{X}$.
Then,
\begin{align}
&\quad ~ |f(\boldsymbol{x}) - \mu_{t-1}(\boldsymbol{x})| \nonumber \\
&= |g(f_1(\boldsymbol{x}), ..., f_J(\boldsymbol{x})) - g(\mu_{1,t-1}(\boldsymbol{x}), ..., \mu_{J,t-1}(\boldsymbol{x}))| \nonumber \\
&\leq \sum\nolimits_{j=1}^J B_j | f_j(\boldsymbol{x}) - \mu_{j,t-1}(\boldsymbol{x}) | \label{eqn:half_normal}
\end{align}

Applying the same argument in Lemma \ref{lemma:chernoff_bound_modified} to function $f_j$:
\begin{align*}
Pr\{|f_j(\boldsymbol{x}) - \mu_{j,t-1}(\boldsymbol{x})| > \beta_t^{1/2} \sigma_{j,t-1}(\boldsymbol{x})\} \leq e^{-\beta_t^{1/2}}
\end{align*}
Then applying the union bound on $j \in [J]$, we get
\begin{align*}
&\quad ~ |f(\boldsymbol{x}) - \mu_{t-1}(\boldsymbol{x})| \\
&\leq \sum\nolimits_{j=1}^J B_j | f_j(\boldsymbol{x}) - \mu_{j,t-1}(\boldsymbol{x}) | \\
&\leq \sum\nolimits_{j=1}^J B_j \beta_t^{1/2} \sigma_{j,t-1}(\boldsymbol{x}) \\
&\leq \beta_t^{1/2} \sqrt[]{J (\sum\nolimits_{j=1}^J B_j^2 \sigma_{j,t-1}^2(\boldsymbol{x}))} \\
&=\beta_t^{1/2} \sigma_{t-1}(\boldsymbol{x})
\end{align*}
with probability $1 - J e^{-\beta_t^{1/2}}$, where the last inequality is from Cauchy's inequality.
By applying union bound again to all $\boldsymbol{x} \in \mathcal{X}$, the above inequality yields:
\begin{align*}
|f(\boldsymbol{x}) - \mu_{t-1}(\boldsymbol{x})| \leq \beta_t^{1/2} \sigma_{t-1}(\boldsymbol{x}) ~\forall \boldsymbol{x} \in \mathcal{X}
\end{align*}
with probability $1 - |\mathcal{X}| J e^{-\beta_t^{1/2}}$.
Choosing $|\mathcal{X}| J e^{-\beta_t^{1/2}} = \delta / \pi_t$ and using the union bound for $t \in \N$, the statement holds, i.e. $\beta_t = 2 \log(|\mathcal{X}| J \pi_t / \delta)$.
Specifically, if we choose $\pi_t = \pi^2 t^2/ 6$, then it implies $\beta_t = 2 \log(|\mathcal{X}| J t^2 \pi^2 / 6 \delta)$.
\qed
\end{proof}

\generalizedDecomposedGPUCBbound*
\begin{proof}
Directly follow by the same proofs of Theorem \ref{thm:decomposed_GPUCB_bound} with Lemma \ref{lemma:regret_variance_bound}, Lemma \ref{lemma:information_gain_modified}, and Lemma \ref{lemma:lemma:chernoff_bound_modified_generalized}.
\qed
\end{proof}

\begin{remark}
In inequality \eqref{eqn:half_normal}, if we write $Z_j = | f_j(\boldsymbol{x}) - \mu_{j,t-1}(\boldsymbol{x}) |$, where $f_j(\boldsymbol{x}) - \mu_{j,t-1}(\boldsymbol{x})$ is sampled from a normal distribution with $0$ mean and $\sigma_{j,t-1}^2(\boldsymbol{x})$ (due to Gaussian process property).
Then this $Z_j$ is a random variable drawn from a half-normal distribution with parameter $\sigma_j(\boldsymbol{x})$ (no longer the variance here).

The summation of half-normal distributions can still be computed and bounded by a similar inequality like inequality \eqref{eqn:chernoff_bound}.
This can provide a constant ratio of improvement to the $\beta_t$ exploration parameter, thus the regret bound as well.
However it does not change the order of regret and sample complexity.
Therefore we are not going to cover this here.
\end{remark}

\end{document}